\UseRawInputEncoding
\pdfoutput=1%
\documentclass[lettersize,journal]{IEEEtran}
\usepackage{amsmath,amsfonts}
\usepackage{algorithmic}
\usepackage{algorithm}
\usepackage{array}
\usepackage[caption=false,font=normalsize,labelfont=sf,textfont=sf]{subfig}
\usepackage{textcomp}
\usepackage{url}
\usepackage{verbatim}
\usepackage{graphicx}
\usepackage{bbding}
\usepackage{pifont}
\usepackage{threeparttable}
\usepackage{makecell}
\usepackage{booktabs} 
\usepackage{graphicx}
\usepackage{float}
\usepackage{subfig}
\usepackage{amssymb}
\usepackage{array}
\usepackage{multirow} 
\usepackage{cite}
\begin{document}

\title{Dual Defense: Adversarial, Traceable, and Invisible Robust Watermarking against
Face Swapping}

\author{Yunming Zhang, Dengpan Ye, Caiyun Xie, Long Tang, Chuanxi Chen, Ziyi Liu, Jiacheng Deng
        % <-this % stops a space
\IEEEcompsocitemizethanks{\IEEEcompsocthanksitem Yunming Zhang, Dengpan Ye, Caiyun Xie, Long Tang, Chuanxi Chen, Ziyi Liu, Jiacheng Deng are with Wuhan University, School of Cyber Science and Engineering, Key Laboratory of Aerospace Information Security and Trusted Computing, Ministry of Education, Wuhan, 430072, China. 
}    
\thanks{}% <-this % stops a space
\thanks{}}

% Journal of \LaTeX\ Class Files,~Vol.~14, No.~8, August~2021
\markboth{}%
{Shell \MakeLowercase{\textit{et al.}}: A Sample Article Using IEEEtran.cls for IEEE Journals}


\maketitle

\begin{abstract}

The malicious applications of deep forgery, represented by face swapping, have introduced security threats such as misinformation dissemination and identity fraud. While some research has proposed the use of robust watermarking methods to trace the copyright of facial images for post-event traceability, these methods cannot effectively prevent the generation of forgeries at the source and curb their dissemination. To address this problem, we propose a novel comprehensive active defense mechanism that combines traceability and adversariality, called Dual Defense. Dual Defense invisibly embeds a single robust watermark within the target face to actively respond to sudden cases of malicious face swapping. It disrupts the output of the face swapping model while maintaining the integrity of watermark information throughout the entire dissemination process. This allows for watermark extraction at any stage of image tracking for traceability. Specifically, we introduce a watermark embedding network based on original-domain feature impersonation attack. This network learns robust adversarial features of target facial images and embeds watermarks, seeking a well-balanced trade-off between watermark invisibility, adversariality, and traceability through perceptual adversarial encoding strategies. Extensive experiments demonstrate that Dual Defense achieves optimal overall defense success rates and exhibits promising universality in anti-face swapping tasks and dataset generalization ability. It maintains impressive adversariality and traceability in both original and robust settings, surpassing current forgery defense methods that possess only one of these capabilities, including CMUA-Watermark, Anti-Forgery, FakeTagger, or PGD methods.

\end{abstract}


\section{Introduction}
\IEEEPARstart{T}{he} groundbreaking research in deep learning has driven the rapid advancement of deep forgery, especially in face swapping techniques, which seamlessly replace the target face with the source face, fabricating identities. Malicious actors can exploit this technology to create high quality images and videos of political figures or celebrities, involving themselves in illegal political and commercial activities, identity fraud, and other unlawful behaviors~\cite{huang2023implicit,chen2022robust}. Therefore, there is an urgent need to implement effective measures to counter the potential threats introduced by face swapping technology.

\begin{figure}[!t]
\centering
\includegraphics[width=1\columnwidth]{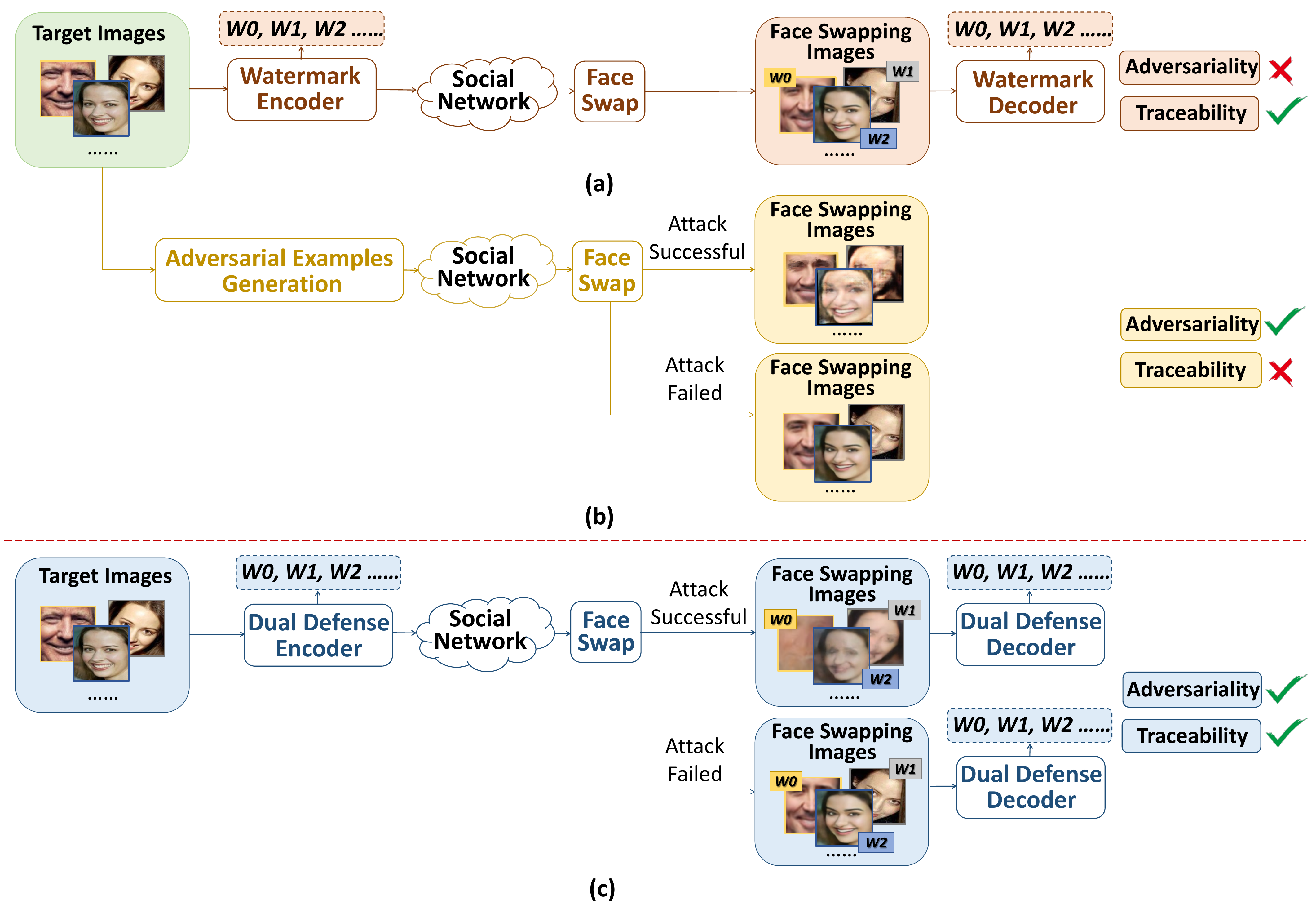}
\caption{Illustration of deep forgery active defense scenarios.  (a) Active defense based on watermarking. Enables tracing the source of forged images but cannot prevent forgery and eliminate its adverse effects at the source. (b) Active defense based on adversarial examples. It can disrupt forgery generation but does not support traceability, offering no traceability basis upon attack failure. (c) Our Dual Defense active defense. While disrupting the output of the model, it can extract the pre-encoded watermark from the disrupted image for image traceability. It also provides auxiliary traceability means for scenarios of attack failure.}
\label{fig_1}
\end{figure}

Existing defenses against deep forgery mainly include passive detection and active defense. Passive detection involves training detectors to retrospectively validate forged images~\cite{jeong2022frepgan,huang2023implicit, wang2023classification}, while active defense methods primarily include deep watermarking-based defense~\cite{wang2021faketagger,yu2021artificial} and adversarial examples-based defense~\cite{huang2021initiative,wang2022anti}, as shown in Fig.~\ref{fig_1}. Deep watermarking algorithm provides active defense by pre-encoding a robust watermark representing identity information into the carrier image. After the image has been forged, users can extract the watermark for tracing its source. However, the robust watermark cannot disrupt the forgery model, thus failing to fundamentally eliminate the impact of malicious forgery. The typical active defense involves generating adversarial perturbations based on gradient iteration and directly overlaying them onto the original pixels. However, its adversariality is significantly reduced after undergoing image post-processing. The adversarial perturbation has a certain vulnerability and cannot be recovered after the carrier is forged, making the adversarial example method non-traceable. In case the adversarial perturbation attack fails on the forgery model, users cannot protect their rights through image traceability.  Therefore, there is an urgent need for a comprehensive active defense method in real-world social networks, capable of disrupting forgery models while tracing the forged results, and withstand most image post-processing operations. 

As a robust copyright protection tool, robust watermarking can be transmitted alongside the carrier in complex channel environments. Therefore, image tracing can be employed in conjunction with adversarial attacks as a joint defense mechanism to ensure the effectiveness of defense methods in any complex scenario. However, the generation method of adversarial examples hinders their simultaneous end-to-end training with robust watermarking models. Additionally, the susceptibility of universal adversarial perturbations makes them unsuitable for direct overlay with watermarks, leading to significant degradation in both adversariality of the adversarial examples and the accuracy of watermark recovery. The characteristics of adversarial examples and robust watermarking have sparked a new consideration for active defense against face swapping. Is it possible to encode the deep watermark into the carrier image in a manner that deviates from the source face manifold, making it adversarial to the face swapping model while maintaining the robustness of the watermark? Building upon this, we focus on face swapping that are more susceptible to legal disputes and propose a novel dual-effect active defense method combating the classic face swapping model FaceSwap, named Dual Defense, which effectively combines adversariality and traceability. 

Dual Defense provides the first validation of the adversariality of invisible robust watermark in the generative model. We introduce a novel adversarial watermarking network based on original-domain feature impersonation attacks. Our research demonstrates that the reconstruction process of the watermarked image of the target face can serve as an imitable feature, effectively circumventing the gradient explosion caused by non-targeted attacks in multi-objective optimization. To achieve this, we specifically design an adversarial loss based on impersonation attacks to embed the watermark into the robust adversarial features of the carrier, causing the output of the FaceSwap model to deviate from the source facial manifold. To alleviate the decline in image quality caused by adversarial optimization, we introduce image quality loss based on structure information compensation. This involves converting images to the low-frequency domain and computing structural similarity to compensate for the loss of semantic information within the pixel domain. Simultaneously, we train the watermark decoder to extract watermark information both before and after face swapping, ensuring the accurate extraction of watermarks at any stage of carrier transmission, catering to diverse tracing requirements.

% Our research demonstrates that the watermark encoder, under the supervision of target facial watermarked images, can learn the robust adversarial features of the carrier. To achieve this, we implement an adversarial impersonation attack on the target face watermarked image through the design of a specialized adversarial loss based on reconstruction feature learning. By embedding the watermark into deep feature maps, it can cause the output of FaceSwap model to deviate from the source facial manifold, achieving adversarial performance for invisible watermarks. Thus, compared to adversarial examples, watermarked images exhibit outstanding adversarial robustness, making them more suitable for proactive defense of facial images in real-world social networks. To alleviate the decline in image quality caused by adversarial optimization, we introduce image quality loss based on structure information compensation. This involves converting images to the low-frequency domain and computing structural similarity to compensate for the loss of semantic information within the pixel domain. Simultaneously, we train the watermark decoder to extract watermark information both before and after face swapping, ensuring the accurate extraction of watermarks at any stage of carrier transmission, catering to diverse tracing requirements.

As shown in Fig.~\ref{fig_1} (c), compared with the existing defense methods, Dual Defense can fundamentally protect the security of face images in social networks. The embedded robust watermark can track the entire transmission process of carrier images within social networks. In the event of malicious face swapping during transmission, Dual Defense can promptly disrupt the output of the face-swapping model while ensuring the integrity of the watermark. Even in complex network environments with multiple image post-processing stages, the watermark maintains a high level of adversarial robustness. Should an attack fail, users can still extract the robust watermark for timely traceability, thereby breaking the transmission chain. Our research potentially opens up a novel research direction for applying deep watermarking in the comprehensive active defense against deep forgery.

Our contributions can be summarized as the following:

\begin{itemize}
% \item A proactive defense method is proposed against the classic face swapping model Faceswap. By embedding a robust adversarial watermark representing identity information into the source face images, it achieves a dual effect of traceability and adversarial defense in deepfake active defense for the first time.
% \item We propose an active defense method targeting the classic face swapping model FaceSwap, called Dual Shield. Dual Shield offers a dual effectiveness in terms of traceability and adversariality and demonstrates a notable level of grey-box transferability and dataset generalization ability.
\item We propose Dual Defense, the first dual-effect active defense method that combines both adversariality and traceability. It exhibits exceptional robustness, cross-task universality and dataset generalization ability.

\item We innovatively present an adversarial watermark network based on original-domain feature impersonation attacks. This network imparts adversariality to invisible robust watermarks and resolves conflicts in multi-objective watermark optimization through a GAN-based perceptual adversarial encoding strategy.
% \item We first present an original-domain facial feature attack method that maintains the robustness of deep watermarking while also making it adversarial to FaceSwap.

% \item We introduce a GAN-based perceptual adversarial encoding strategy, which minimizes the image quality degradation caused by adversarial enhancement while achieving a trade-off among multiple performance.
% \item We innovatively introduce a GAN-based perceptual adversarial encoding strategy, which resolves conflicts in multi-objective optimization of watermarking and achieves a balance among multiple performance aspects.

% \item We specifically design a more reasonable evaluation method to fully assess the adversarial effects in face swapping tasks. Additionally, we propose a comprehensiveactive defense evaluation metric to holistically evaluate defense effectiveness from both adversariality and traceability perspectives.

\item We specifically design a more reasonable and comprehensive evaluation method to fully assess the adversariality of the proposed Dual Defense against face swapping.

% \item Extensive experiments and visualizations demonstrate the superiority of our method over the state-of-the-art approaches.

\item Extensive experiments demonstrate the superiority of Dual Defense. Its adversarial capabilities reduce the face recognition success rate to below 0.004, while achieving a 0.99 accuracy in recovering watermark information.
\end{itemize}

\section{Related Works}

\subsection{Deep Forgery Models}

Deep forgery models can be categorized into four types based on their task objectives: attribute editing~\cite{Choi_Choi_Kim_Ha_Kim_Choo_2018,Hall_Xu_Sebe_Yan_2019}, face reenactment~\cite{Hong_2022_CVPR}, full-face synthesis~\cite{karras2019style, shen2020interpreting} and face swapping~\cite{chen2020simswap,natsume2018rsgan,natsume2019fsnet,korshunova2017fast}. Differing from the other three types of manipulations, face swapping techniques involve changes to personal identity information, which makes them more prone to legal issues related to facial privacy and image copyrights. So we focus on face swapping methods. FaceSwap~\cite{Faceswap} is a representative algorithm in the field of deep forgery technology, which uses an autoencoder to achieve face swapping between two individuals. GAN-based face swapping methods, such as FSGAN~\cite{2019fsgan} and FaceShifter~\cite{2020advancing}, improve the ability to represent facial features and generate controllable results, leading to higher quality generations. The AOT algorithm ~\cite{2020aot} addresses the chromatic attribute differences in the replacement process from the perspective of optimal transport. Among these methods, the autoencoder structure of FaceSwap significantly improves the operability of face swapping while also lowering the technical threshold, making it the most widely used face swapping algorithm to date.

 \subsection{Robust Watermarking-based Active Defense}
Robust watermarking technology plays a vital role in the field of information hiding~\cite{zhang2023towards,zhang2021jnd}. Recent research has focused on using invisible watermarks for deep forgery active defense. Wang et al.~\cite{wang2021faketagger} utilize the deep watermarking method to encode watermarks into carrier face images, enabling effective tracing of manipulated face images. Yu et al.~\cite{yu2021artificial} introduce Artificial Fingerprints, validating their transferability from training data to generative models. However, the aforementioned active defense method based on robust watermarking lacks adversariality. In cases of sudden malicious forgery on the carrier, the watermark is unable to disrupt the forgery model and mitigate its adverse impact. Recent studies propose applying visible watermark patches to attack classification models~\cite{jia2020adv, jiang2021fawa}. However, visible watermarks can affect the normal use of images and are prone to detection and removal by attackers. In this paper, we investigate the adversarial robustness of robust watermarks against face-swapping models, enabling them to actively disrupt the generation model while tracking carrier copyrights, thus fundamentally preventing malicious events.

\subsection{Adversarial Examples-based Active Defense}
% Adversarial examples were initially applied to image classification tasks~\cite{croce2020reliable, 10.1145/3503161.3547989}, and in recent years, they have gradually extended to generative models~\cite{huang2021initiative}. ~\cite{yeh2020} propose Distorting Attack, utilizing PGD to target the generator and perform white-box attacks on attribute editing models. ~\cite{wang2022anti} introduce Anti-Forgery, an active defense method against deepfakes using robust adversarial perturbations. This approach is applicable to various image transformation scenarios. To achieve cross-model image-agnostic adversarial perturbations, ~\cite{huang2022cmua} propose CMUA-Watermark, which solves the problem of mutual cancellation of anti-noise between different images and different models. ~\cite{yang2021defending} propose Adversarial Faces, a PGD-based adversarial attack against FaceSwap training, resulting in low-quality, easily detectable face swapping images. However, real-time application is not feasible, limiting its suitability for social network users. Adversarial examples are not traceable and cannot be applied to copyright protection scenarios.

The DeepFake disruption is the promising countermeasures for fighting against DeepFakes in a proactive manner~\cite{chen2021magdr}. Yeh et al.~\cite{yeh2020} propose Distorting Attack, utilizing PGD to target the generator and perform white-box attacks on attribute editing models. To achieve cross-model image-agnostic adversarial perturbations, Huang et al.~\cite{huang2022cmua} propose CMUA-Watermark, which solves the problem of mutual cancellation of anti-noise between different images and different models. Li et al.~\cite{li2022unganable} propose UnGANable, which attacks GAN-inversion based face manipulation by searching for alternative images around the original images in image space. But the above methods are only effective in attribute editing and face reenactment since these two types of manipulation model share a similar pipeline~\cite{wang2022anti}. The face swapping model modifies the high-dimensional semantic features of the images, which is more challenging to defend than the attribute editing model. Wang et al.~\cite{wang2022anti} introduce Anti-Forgery, an active defense method against deepfakes using robust adversarial perturbations. This approach is applicable to various image transformation scenarios. However, it requires multiple iterations for each image, resulting in noticeable degradation of perceptual quality and showing poor robustness to common image processing operations. Yang et al. ~\cite{yang2021defending} propose Adversarial Faces, a PGD-based adversarial attack against FaceSwap training, resulting in low-quality, easily detectable face swapping images. However, real-time application is not feasible, limiting its suitability for social network users. Current adversarial example-based active defense methods exhibit defense vulnerabilities due to the fragility of adversarial perturbations. They are susceptible to attack failures when adversarial examples undergo post-processing operations, rendering users incapable of preventing malicious events using auxiliary means. Therefore, in this paper, we investigate the robust adversariality of traceable watermarks in complex real-world network environments. The proposed Dual Defense solution also offers complementary measures for worst cases of attack failure. 

We present a comparative analysis of the proposed method with several existing representative active defense methods in Table~\ref{Func com}. The data embedded in the carrier can be categorized into two types: perturbations, generated through iterative adversarial training and irrecoverable, and watermarks, which are user-customizable, contain copyright information, and can be recovered using a decoder for carrier tracing.

\begin{table}[t]
\caption{Comparison of the relevant active defense methods.}
\centering
	
	% \renewcommand\arraystretch{1}  % 调整行高
	% \scriptsize  % 字号
 % \setlength\tabcolsep{6.7pt} 
	% \resizebox{.95\linewidth}{1} % 这句总是报错
 % \scalebox{0.2} % dz换了一个表格缩放语句
 \resizebox{\hsize}{!}{
\begin{tabular}{ccccccccc}

			\hline
			\multirow{2}{*}{\textbf{Method}} &\multirow{2}{*}{\textbf{Type}} & \multicolumn{2}{c}{\textbf{Adversariality}} & \multicolumn{2}{c}{\textbf{Traceability}}\\ \cline{3-4} \cline{5-6}
   
                & & Original & Robust & Original & Robust\\
                 
			%  \hline
			%  \multicolumn{9}{c}{SST}\\
			\hline
			CMUA-Watermark~\cite{huang2022cmua} & Perturbation&  \Checkmark & \ding{56} &\ding{56} &\ding{56}\\
                Distorting Attack~\cite{yeh2020} & Perturbation&  \Checkmark & \ding{56} &\ding{56} &\ding{56}\\
              Anti-Forgery~\cite{wang2022anti} & Perturbation& \Checkmark & \Checkmark &\ding{56} &\ding{56}\\
              Adversarial Faces~\cite{yang2021defending} & Perturbation& \Checkmark & \Checkmark &\ding{56} &\ding{56}\\
             Artificial Fingerprints~\cite{yu2021artificial} & Watermark& \ding{56} & \ding{56} &\Checkmark &\Checkmark \\
		FakeTagger~\cite{wang2021faketagger} & Watermark& \ding{56} & \ding{56} &\Checkmark &\Checkmark\\

			  \textbf{Dual Defense (ours)} & Watermark & \Checkmark & \Checkmark &\Checkmark &\Checkmark\\
			\hline
\end{tabular}}

\label{Func com}
\end{table}

\begin{figure}
    \centering
    \includegraphics[width=75mm]{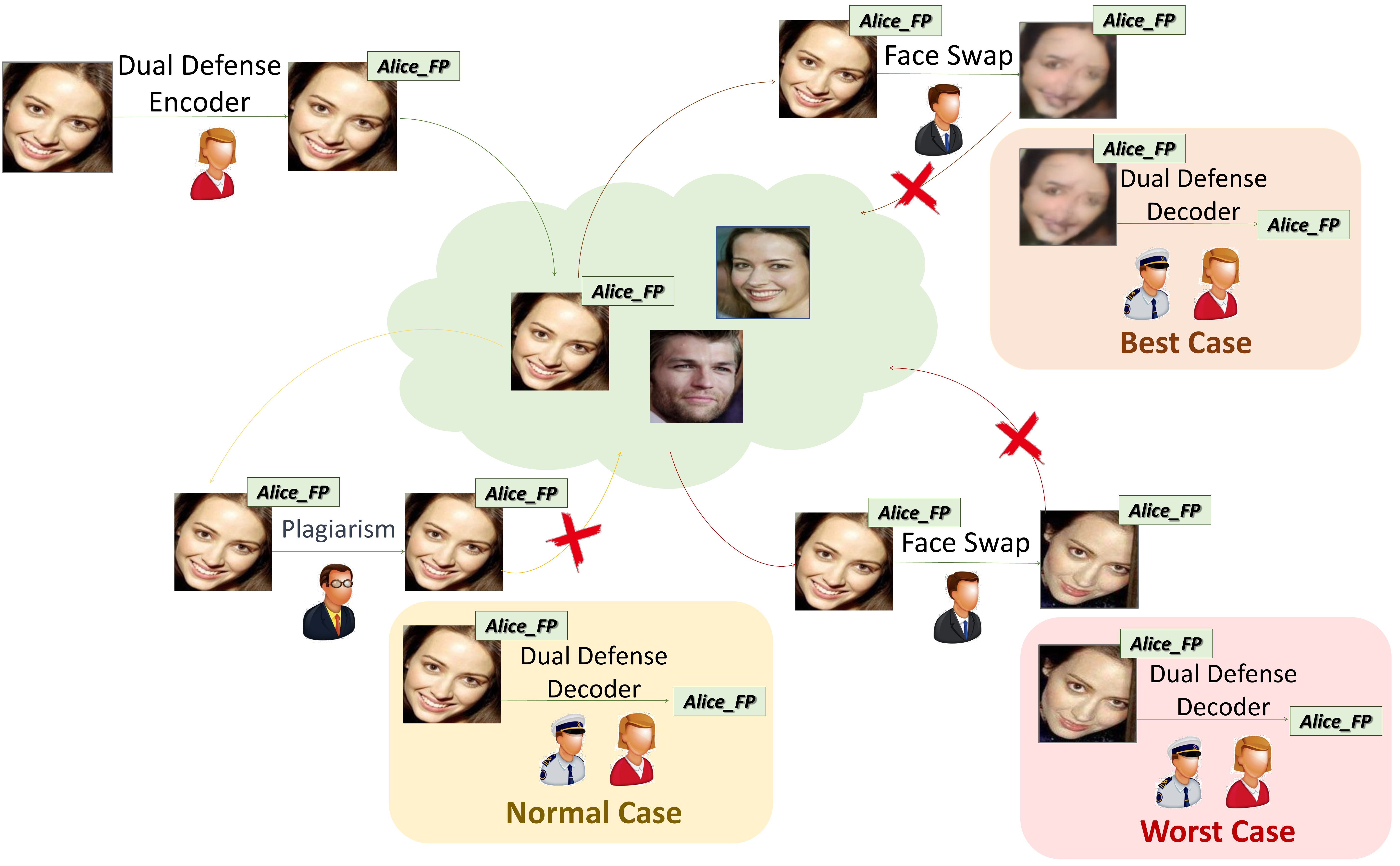}
    \caption{Explanatory example of Dual Defense for active defense. Defensive measures are presented for three different threat cases. \textbf{Norm case}: Standard copyright tracking scenario to prevent plagiarism and theft. \textbf{Best case}: The most robust defense attack scenario where we can extract the watermark for tracing even while defending against face-swap models. \textbf{Worst case}: Defense scenario when adversarial attacks fail due to excessive image post-processing, and we can prevent further propagation by extracting the watermark for tracing.}
    \label{sn}
\end{figure}
    
\section{Problem Statement}

\label{PS}
In this section, we present the problem definition of our Dual Defense. Given the target user $L_{t}$, who possesses a personal image collection $\{ x_{i} \}_{i=1}^{N} \in X_{t}$ containing $N$ face images $x_{i}$. In the Dual Defense protection mechanism, users can pre-embed watermark into images using the encoder $En ( \cdot )$. This watermark is user-defined and can include, but is not limited to, IDs, timestamps, or scene markers. Serving as a form of generative AI fingerprint, the watermark is rooted within the deep image features to trace image copyrights:

\begin{eqnarray}
    X_{t} ^{W_{ID}}=En(X_{t},  W_{ID}),
\end{eqnarray}
where $X_{t} ^{W_{ID}}$ represents the obtained watermarked image set $\{x_{i}^{W_{ID} } \}_{i=1}^{N}   \in X_{t}^{W_{ID}}$. Users can then upload the watermarked images to social networks. Based on the context of a user's facial image transmission on social networks, as shown in Fig.~\ref{sn}, Dual Defense offers proactive defense strategies from three key cases, encompassing a wide range of potential scenarios:

\textbf{Normal case:} First and foremost, Dual Defense, as a robust watermarking method, possesses the fundamental capability for copyright tracking. The custom watermark can trace the entire dissemination process of the watermarked image. In cases where there are no sudden instances of malicious face swapping, the watermarked image may still be vulnerable to illegal activities such as plagiarism and unauthorized distribution. 

When watermarked images face copyright disputes, users can employ the watermark decoder $De(\cdot )$ to extract the pre-embedded watermark representing copyright ownership for tracing the source:
 \begin{eqnarray}
    W_{norm}= De(SN( X_{t} ^{W_{ID}})),
\end{eqnarray}
where $SN(\cdot)$ represents social networks, and $W_{norm}$ represents the extracted watermark information in this scenario.

\textbf{Best case:} 
During the distribution of watermarked images, copyright is persistently at risk, particularly when unauthorized individuals exploit them for malicious face swapping. Traditional robust watermarking methods can merely track and provide evidence post-incident, falling short of genuinely thwarting such misconduct. However, Dual Defense introduces strong adversarial capabilities in these scenarios. It can disrupt the outputs of the FaceSwap model while preserving the integrity of robust watermarks, thus guaranteeing end-to-end copyright tracking.

When watermarked images are subjected to malicious face swapping, the malicious forger can use FaceSwap model $FS(\cdot)$ to perform face swapping with the source face images $\{ x_{m}  \}_{m=1}^{M} \in X_{s}$ that belongs to the source user  $L_{s}$. The adversariality of the watermark can disrupt the output of FaceSwap model, resulting in visually distorted image dataset $X_{adv}^{W_{ID}}$ where identity is unrecognizable:

\begin{eqnarray}
   X_{adv}^{W_{ID}}=FS(X_{t}^{W_{ID}},'{L_{s}}').
   \label{advfs}
\end{eqnarray}

 However, this image may still retain background information related to the original identity, making it susceptible to secondary manipulation for the dissemination of fake news, which poses a significant threat, especially to public figures. In such cases, the original user can trace the source and provide evidence of image ownership by extracting the pre-embedded watermark information representing a timestamp or original scene tag:
 
\begin{eqnarray}
W_{best}= De(X_{adv}^{W_{ID}}).
\label{advw}
\end{eqnarray}

Where $W_{best}$ represents the extracted watermark information in this best case.

\textbf{Worst case:} During image transmission on social networks, various post-processing operations~\cite{9103635}, including compression, noise, and filtering, are applied. Extensive and complex post-processing can render some adversarial methods nearly useless. While Dual Defense maintains robust adversarial capabilities and can resist most image processing operations, there are still rare extreme cases that can cause attacks to fail, enabling FaceSwap models to successfully generate swapped images. In such case, network administrators or the original users can detect the true ownership of the images by extracting the watermark, promptly halting the dissemination of forged images, and preventing the spread of detrimental consequences.

% When watermarked images undergo transmission in complex environments and are subjected to malicious face swapping, the adversariality of the watermark may decrease, potentially resulting in the successful alteration of some images, yielding face-swapped watermark image $X_{s(t)}^{W_{ID}}$. In such cases, network administrators or the original users can detect the true ownership of the images by extracting the watermark, promptly halting the dissemination of forged images, and preventing the spread of detrimental consequences.
\begin{eqnarray}
   X_{s(t)}^{W_{ID}}=FS(X_{t}^{W_{ID}},'{L_{s}}'),
   \label{adv2}
\end{eqnarray}

\begin{eqnarray}
W_{worst}= De(X_{s(t)}^{W_{ID}}),
\label{adv3}
\end{eqnarray}
where $X_{s(t)}^{W_{ID}}$ represents face swapped watermarked image set, and $W_{worst}$ represents the extracted watermark information in this scenario.
\begin{figure*}
	\centering
	\includegraphics[width=155mm]{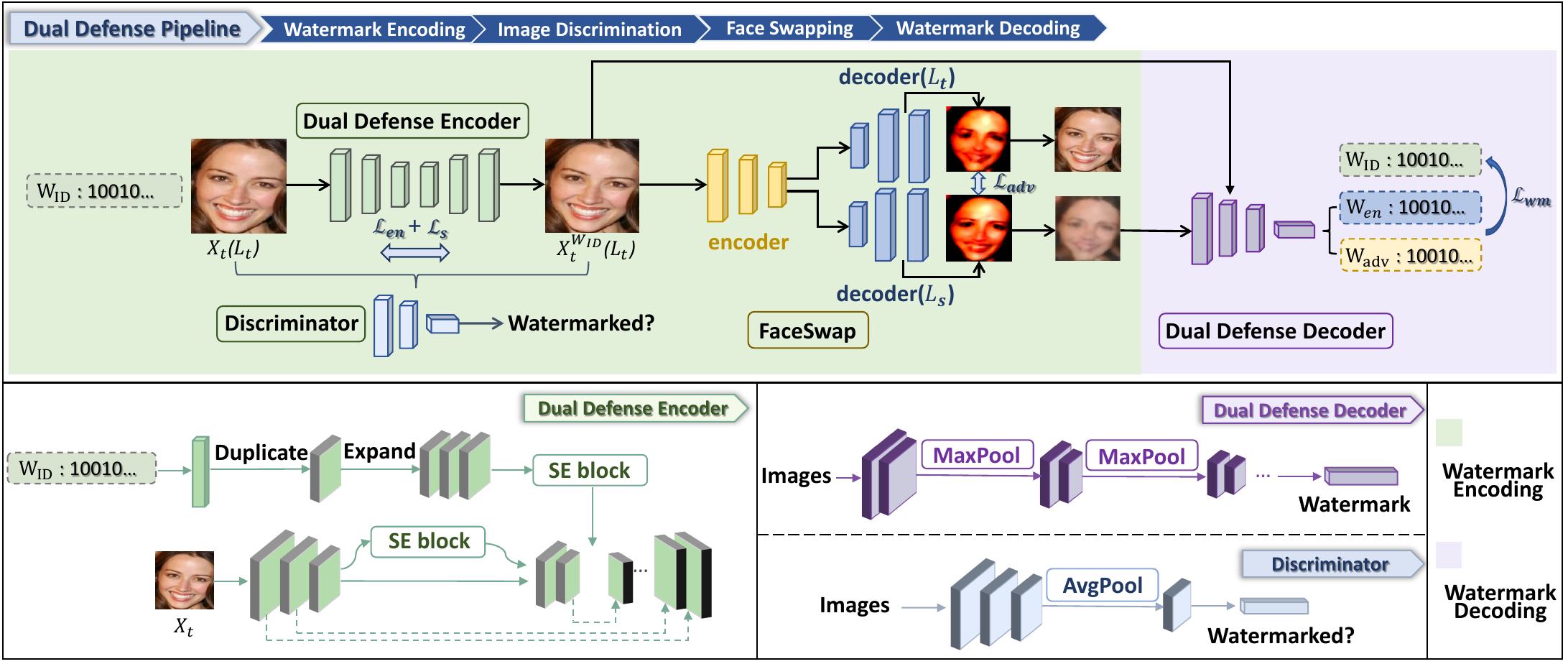}
        \caption{The whole pipeline of Dual Defense. Dual Defense optimizes watermarking model through end-to-end training. The process begins by inputting target images $X_{t}$, along with the user-defined watermark $W_{ID}$, into the encoder to generate watermarked images. Subsequently, the watermarked images undergo FaceSwap, followed by the original-domain feature impersonation attacks. Both disturbed images and watermarked images then pass through the watermark decoder for decoder optimization.}	
 \label{TARA-Watermark outline}
\end{figure*}

\begin{algorithm}[tb]
    \caption{Dual Defense Training Framework}
    \label{alg:TARA-Watermark}
    \textbf{Input}: $X_{t}$ (training original target images), $W_{ID}$ (watermarks), $En( \cdot)$ (watermark encoder), $De( \cdot )$ (watermark decoder), $FS( \cdot )$ (FaceSwap model), $maxiter$ (number of iterations), $deiter$ (number of iterations when the decoder starts to update).\\
    \textbf{Output}: best model parameters.
    \begin{algorithmic}[1]
        \STATE  Initialization;    
        \FOR{$i\in  [ 0,maxiter  ]$ }
           \STATE $X_{t}^{W_{ID}}  \Leftarrow En(X_{t}, W_{ID}) $;
           % \STATE $0 or 1  \Leftarrow Di\left (X_{t}, X_{t }^{W_{ID}} \right ) $;
           \STATE $X_{t }^{(t)}  \Leftarrow FS(X_{t }^{W_{ID}},'{L_{t}}' ) $;
           
           \STATE $X_{adv}^{W_{ID} }  \Leftarrow FS(X_{t }^{W_{ID}},'{L_{s}}') $;
            \STATE 0 or 1$\Leftarrow Di( X_{t}^{W_{ID} }, X_{t} ) $;
           \STATE Compute disciminator loss $\mathcal{L}_{D}$ with Eq.(\ref{Dloss});
           \STATE Update disciminator;
           \STATE Compute image loss $\mathcal{L}_{img}$ with Eq.(\ref{enloss}--\ref{imgloss});
            \IF {$i>deiter$}
                \STATE $W_{en},W_{adv}\Leftarrow  De(X_{t}^{W_{ID} }, X_{adv}^{W_{ID} })$;
                \STATE  Compute message loss $\mathcal{L}_{wm}$ with Eq.(\ref{wmenloss}--\ref{wmloss})
                \STATE $\mathcal{L}_{total}=\alpha \mathcal{L}_{img}+\beta \mathcal{L}_{wm}$;
                \STATE Update watermark encoder and decoder;
            \ELSE
                \STATE $\mathcal{L}_{total} = \mathcal{L}_{img}$;
                \STATE Update watermark encoder;
            \ENDIF
        
        \ENDFOR
        
        \RETURN best model parameters.
    \end{algorithmic}
\end{algorithm}

% \begin{algorithm}[H]
% \caption{Dual Defense Training Framework.}\label{alg:alg1}
% \begin{algorithmic}
% \STATE 
% \STATE {\textsc{TRAIN}}$(\mathbf{X} \mathbf{T})$
% \STATE \hspace{0.5cm}$ \textbf{select randomly } W \subset \mathbf{X}  $
% \STATE \hspace{0.5cm}$ N_\mathbf{t} \gets | \{ i : \mathbf{t}_i = \mathbf{t} \} | $ \textbf{ for } $ \mathbf{t}= -1,+1 $
% \STATE \hspace{0.5cm}$ B_i \gets \sqrt{ \textsc{max}(N_{-1},N_{+1}) / N_{\mathbf{t}_i} } $ \textbf{ for } $ i = 1,...,N $
% \STATE \hspace{0.5cm}$ \hat{\mathbf{H}} \gets  B \cdot (\mathbf{X}^T\textbf{W})/( \mathbb{1}\mathbf{X} + \mathbb{1}\textbf{W} - \mathbf{X}^T\textbf{W} ) $
% \STATE \hspace{0.5cm}$ \beta \gets \left ( I/C + \hat{\mathbf{H}}^T\hat{\mathbf{H}} \right )^{-1}(\hat{\mathbf{H}}^T B\cdot \mathbf{T})  $
% \STATE \hspace{0.5cm}\textbf{return}  $\textbf{W},  \beta $
% \STATE 
% \STATE {\textsc{PREDICT}}$(\mathbf{X} )$
% \STATE \hspace{0.5cm}$ \mathbf{H} \gets  (\mathbf{X}^T\textbf{W} )/( \mathbb{1}\mathbf{X}  + \mathbb{1}\textbf{W}- \mathbf{X}^T\textbf{W}  ) $
% \STATE \hspace{0.5cm}\textbf{return}  $\textsc{sign}( \mathbf{H} \beta )$
% \end{algorithmic}
% \label{alg1}
% \end{algorithm}

\section{Method}

In this section, we present a detailed explanation of the proposed Dual Defense pipeline, which comprises four components: watermark encoder, discriminator, FaceSwap model, and watermark decoder, as shown in Fig.~\ref{TARA-Watermark outline}. The multiple optimization objectives of watermark invisibility, adversariality, and traceability pose a difficult trade-off. One of the main challenges that Dual Defense aims to address is how to encode watermarks into the original target images in a way that is more perceptually aligned with human vision while ensuring adversariality and traceability, balancing the optimization conflicts among different watermark performance aspects. 

\subsection{Dual Defense Encoder}

In the optimization process of the watermark encoder, we propose a perceptual adversarial encoding strategy based on original-domain feature impersonation attack. This strategy embeds the watermark into the carrier's robust feature maps in a manner that deviates from the source facial manifold. During watermark embedding, the watermark message is duplicated and expanded to match the channel number and size as the carrier image feature map after down sampling. It is then concatenated with the feature map for subsequent feature encoding. Therefore, we need to account for not only the influence of spatial information on watermark encoding but also the inter-channel relationships. To achieve this, we introduce SENet into the encoding process~\cite{hu2018squeeze}. SENet adaptively recalibrates channel-wise feature responses by explicitly modelling interdependencies between channels to determine the optimal watermark embedding strength. Furthermore, we integrate a CNN-based discriminator into our framework. The discriminator is trained to differentiate between watermarked images and carrier images, enabling cooperative optimization with the watermark encoder to enhance the quality of watermarked images. The watermark encoder synergistically optimizes both invisibility and adversarial aspects.
% In the watermark encoding process, we propose a GAN-based perceptual adversarial encoding strategy to effectively balance various aspects of watermark performance. In our approach, the watermark message is duplicated and expanded to match the channel number and size as the carrier image feature map after down sampling. It is then concatenated with the feature map for subsequent feature encoding. Therefore, we need to account for not only the influence of spatial information on watermark encoding but also the inter-channel relationships.  We introduce CBAM ~\cite{woo2018cbam} to emphasize important feature regions and determine the optimal watermark encoding strength.The channel attention maps quantify the degree of attention allocated to different channels during the optimization processes of image quality and adversariality. By incorporating the channel attention mechanism, the watermark encoder can identify the optimal strength for watermark encoding. Furthermore, in order to minimize the degradation of watermarked images quality resulting from adversariality optimization, we integrate a CNN-based discriminator into our framework. The discriminator is trained to differentiate between watermarked images and carrier images, enabling cooperative optimization with the watermark encoder to enhance the quality of watermarked images.

\subsubsection{Invisibility Optimization} Adversarial embedding method comes at the cost of significantly degrading the perceptual quality of the watermarked image. Low-frequency information in an image encompasses its brightness and structural semantic features, whereas high-frequency information comprises numerous texture details~\cite{yang2021real}. Traditional image reconstruction objective functions, like the MSE function, have typically prioritized the restoration of global brightness and color while overlooking structural details such as edges and semantic information. We aim to minimize the modification of the carrier image's semantic information and preserve its structural features during the watermark embedding learning process. To achieve this, we introduce image structural information compensation to effectively mitigate the decline in carrier image quality caused by adversarial processing. The invisibility loss of the watermarked image consists of three parts: discriminator loss $\mathcal{L}_{D}$,  image quality loss $\mathcal{L}_{en}$ and structural information compensation loss $\mathcal{L}_{s}$. We first use Binary Cross Entropy (BCE) loss $\mathcal{L}_{D}$to update the discriminator $Di( \cdot)$:

\begin{eqnarray}\label{Dloss}
\begin{split}
\mathcal{L}_{D}&=\mathbb{E} _{x_{i}\sim X_{t}} -log(Di(x_{i}))\\
     &+ \mathbb{E} _{x_{i}\sim X_{t}} -log[1-Di(En(x_{i} ))].
\end{split}
\end{eqnarray}

MSE loss $\mathcal{L}_{en}$ is used to constrain the image distortion after watermark encoding.

\begin{eqnarray}\label{enloss}
\mathcal{L} _{en}=\frac{1}{N}\sum_{i=1}^{N}(x_{i}-En(x_{i} ) ) ^{2}.
\end{eqnarray}

Due to the limitations of the MSE function, we further enhance structural information compensation in the low-frequency part of the image. We convert both the original and watermarked images to the YCbCr color space, where the Y channel contains most of the texture information of the images. Therefore, we perform Discrete Wavelet Transform (DWT) on the Y channel of the image and extract its low-frequency sub-band: 

\begin{eqnarray}\label{DWT}
LL,LH,HL,HH = DWT(Y_{x_{i}}),
\end{eqnarray}
\begin{eqnarray}\label{DWT}
LL',LH',HL',HH' = DWT(Y_{En(x_{i})}),
\end{eqnarray}
where $Y_{x_{i}}$ represents the $Y$ channel of the original carrier image, $Y_{En(x_{i})}$ represents the $Y$ channel of the watermarked image, $LL$ and $LL'$ respectively represent the low-frequency subbands of the original image and watermarked image, while $LH$, $HL$, $HH$ and $LH'$, $HL'$, $HH'$ represent the high-frequency subbands in different directions of the original image and watermarked image. We employ the Structural Similarity (SSIM) loss~\cite{Wang_Bovik_Sheikh_Simoncelli_2004} $\mathcal{L} _{s}$ to measure the structural information difference between the original carrier and the watermarked image in the low frequency subband, serving as compensation for invisibility loss:
\begin{eqnarray}\label{ssim}
\mathcal{L} _{s}= \mathcal{L} _{ssim}=1-SSIM(LL,LL_{en}).
\end{eqnarray}

% \begin{eqnarray}\label{DWT_EN}
% LL',LH',HL',HH' = DWT(En(x_{i} ))
% \end{eqnarray}

% We use the MSE loss to constrain the image distortion after watermark encoding and the BCE loss to optimize the discriminator $Di( \cdot)$. The invisibility loss of the watermarked image consists of three parts: discriminator loss $\mathcal{L}_{D}$, generator loss $\mathcal{L}_{G}$, and image quality loss $\mathcal{L}_{en}$. These are as follows:

% \begin{eqnarray}\label{Dloss}
% \begin{split}
% \mathcal{L}_{D}&=\mathbb{E} _{x_{i}\sim X_{t}} -log(Di(x_{i}))\\
%      &+ \mathbb{E} _{x_{i}\sim X_{t}} -log[1-Di(En(x_{i} ))], 
% \end{split}
% \end{eqnarray}
% \begin{eqnarray}\label{Gloss}
% \mathcal{L}_{G}=\mathbb{E} _{x_{i}\sim X_{t}} log( 1-Di(En(x_{i}))), 
% \end{eqnarray}
% \begin{eqnarray}\label{enloss}
% \mathcal{L} _{en}=\frac{1}{N}\sum_{i=1}^{N}(x_{i}-En(x_{i} ) ) ^{2}.
% \end{eqnarray}
\subsubsection{Adversariality Optimization.}
FaceSwap achieves face swapping by exchanging the decoders corresponding to different individuals. For target face images, FaceSwap employs the encoder $FS_{E}(\cdot )$ to extract the target face features and utilizes the source face decoder $FS_{D}^{s}(\cdot )$ for face swapping:
\begin{eqnarray}
X_{s}^{(t)}= FS_{D}^{s} (FS_{E}(X_{t}) ).
\end{eqnarray}
Dual Defense uses the original-domain facial features of watermarked images as the optimization target, allowing the face swapping process to learn the attribute features during the watermark image reconstruction, thus attacking the FaceSwap decoder. We first reconstruct the encoded features of the target watermarked image using the corresponding target face decoder $FS_{D}^{t}(\cdot )$, extract the output feature map $I_{i(t)}^{W_{ID} }$of the hidden layer of the decoder as the original-domain attack object. We then use the source face decoder to decode the target feature map, obtaining the face swapped feature map $I_{i(s)}^{W_{ID} }$:
\begin{eqnarray}
I_{i(t)}^{W_{ID} } =FS_{D}^{t}(FS_{E}(x_{i}^{W_{ID}} ))_{k-1},  
\end{eqnarray}
\begin{eqnarray}
I_{i(s)}^{W_{ID} } =FS_{D}^{s}(FS_{E}(x_{i}^{W_{ID}} ))_{k-1},  
\end{eqnarray}
where $k$ represents the last layer in the FaceSwap decoder, which is the sigmoid layer. The sigmoid function maps the feature map data to the $[0,1]$ space, narrowing the search space for gradient optimization, so we perform targeted attack by extracting the input feature map of the sigmoid. The adversariality loss function is defined as follows: 
\begin{eqnarray}\label{advloss}
\mathcal{L} _{adv}=\frac{1}{N} \sum_{i=1}^{N} (I_{i(t)}^{W_{ID} }-I_{i(s)}^{W_{ID} } )^{2} .
\end{eqnarray}

\begin{figure}
	\centering
	\includegraphics[width=0.95\columnwidth]{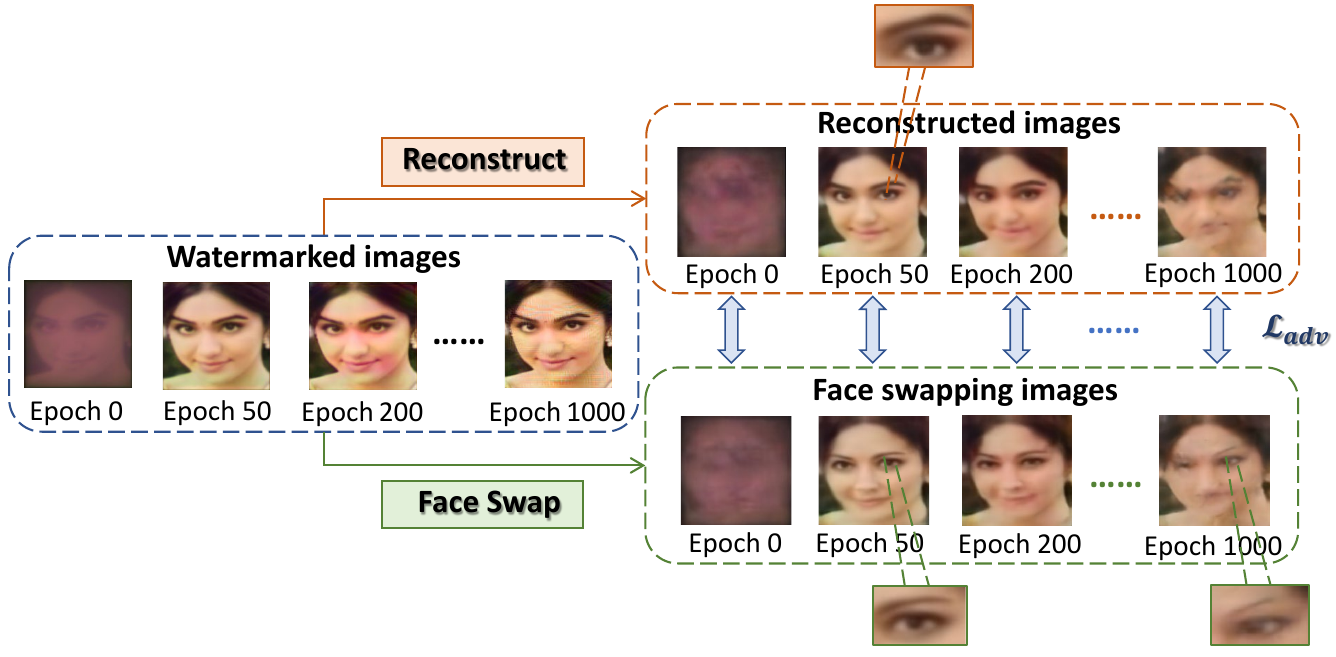}

        \caption{Iterative optimization process of watermarked image adversarial performance. \emph{Reconstruction} refers to using the target face decoder for reconstruction. \emph{Face Swap} indicates using the source face decoder for face swapping.}	
 \label{adv}
\end{figure}

Dual Defense enhances various performances through multi-objective optimization. In the initial iteration, the watermark encoder introduces significant distortion to the watermarked image, leading to severe distortion in the FaceSwap output. Traditional untargeted attacks may cause gradient explosion, hindering multi-objective optimization. Therefore, Dual Defense employs target face reconstruction features for targeted attacks, avoiding gradient explosion while preserving the identity characteristics of the target face simultaneously. Additionally, the attack targets the reconstructed image of the watermarked image instead of the original image. This gradual optimization process allows the encoding learning of the watermark to progress alongside image quality optimization, preventing failures in watermark extraction caused by large disparities in image quality during the initial iterations. Fig.~\ref{adv} shows the watermarked images, their reconstructed images, and the face swapping images at different epochs. It is evident that face swapping images gradually align with reconstructed images of watermarked images in each epoch. At about 50 epochs, a relatively successful face swapping outcome is achieved. However, in subsequent epochs, interference from the target face features further disrupts the face swapping output, revealing certain features of the original target face, such as eyebrows.

The image loss of Dual Defense is the weighted sum of invisibility loss and adversariality loss:
\begin{eqnarray}\label{imgloss}
\mathcal{L}_{img}=\lambda _{en}\mathcal{L}_{en}+\lambda _{s} \mathcal{L}_{s}+ \lambda _{adv} \mathcal{L}_{adv}, 
\end{eqnarray}
where $\lambda _{en}$, $\lambda _{s}$ and $ \lambda _{adv}$ are empirically set to 0.8, 0.1, 0.1 respectively.

\subsection{Dual Defense Decoder}

The watermark decoder needs to decode watermark from both the watermarked image and the disrupted image after face swapping, this is done to avoid overfitting to a specific type of attack and to meet the traceability requirements in various real-world scenarios. We utilize stacked convolutional layers architecture for the decoder and finally employ a max-pooling layer and a linear layer to map the high-dimensional feature back to the original watermark size. The watermark decoder often improves robustness by introducing a noise pool for watermark decoding. However, since Dual Defense includes the FaceSwap process, which already involves significant modifications to image features, there is no need to introduce additional noise during training.

The traceability loss consists of watermarked image message loss $\mathcal{L}_{wm\_en}$ and disrupted image message loss $\mathcal{L}_{wm\_adv}$. The distance between extracted watermark message and original watermark message $W_{ID}$ is measured by the BCE loss  $\mathcal{L }_{BCE}$: 

\begin{eqnarray}\label{wmenloss}\begin{aligned}
   \mathcal{L}_{wm\_en} = \mathcal{L} _{BCE}( W_{ID},W_{en}) ,
\end{aligned}
\end{eqnarray}
\begin{eqnarray}\label{wmadvloss}\begin{aligned}
\mathcal{L}_{wm\_adv}  =\mathcal{L} _{BCE}( W_{ID},W_{adv}),
\end{aligned}
\end{eqnarray}
\begin{eqnarray}\label{wmloss}
\mathcal{L}_{wm}=\mathcal{L}_{wm\_en}+\mathcal{L}_{wm\_adv}.
\end{eqnarray}
% where $W_{en}$ represents the watermark message recovered from the watermarked image. $W_{adv}$ denotes the watermark message recovered from the disrupted image. 

\begin{table*} [t]
\caption{Quantitative results of Dual Defense in original settings without image post-processing.  * represents across different FaceSwap tasks within the same dataset.}
	% \renewcommand\arraystretch{2}
	% \newcommand{\tabincell}[2]
	% \footnotesize
	\centering

	\renewcommand\arraystretch{1.2}  % 调整行高
    % \scalebox{0.88} % dz换了一个表格缩放语句
\resizebox{\hsize}{!}{
   % \begin{center}
       \begin{tabular}{ccccccccccccc}
			\hline
            %     &Method & Adversariality &Traceability\\
 		\multirow{2}{*}{\textbf{Train}}&\multirow{2}{*}{\textbf{Test}} & \multicolumn{2}{c}{\textbf{Invisibility}}& \multicolumn{6}{c}{\textbf{Adversariality}}  & \multicolumn{2}{c}{\textbf{Traceability}}\\
            \cmidrule(r){3-4}\cmidrule(r){5-10} \cmidrule(r){11-12} 
   
           & &PSNR$\uparrow$ & SSIM$\uparrow$  &PSNR$\downarrow$ & SSIM$\downarrow$ & $L_{1}$$\uparrow$ & LPIPS$\uparrow$ & $FN_{acc}$ $\downarrow$& $SR_{mask}$ $\uparrow$&$Acc_{org}$$\uparrow$ &$Acc_{adv}$$\uparrow$\\
			\hline
   
			\multirow{3}{*}{\makecell{VGG-\\Face2}} &VGGFace2 &31.782  &0.917  &20.368  &0.631 &0.078  &0.312 &0.003 &0.772 &0.996 &0.923 \\
                                        &VGGFace2* &30.121 &0.878 & 23.703    &0.833 &0.057 & 0.254 &0.197 &0.612 &0.957 &0.868 \\
			                         &CASIA-WebFace &30.021 &0.849 &22.348     &0.791  &0.055 &0.222 &0.027 &0.582 &0.974 &0.843\\
			
			\hline
   			\multirow{3}{*}{\makecell{CASIA-\\WebFace}} &CASIA-WebFace &31.830  &0.925  & 22.363  &0.764 &0.062  &0.273 &0.009 &0.683   &0.998 &0.986 \\
                                       &CASIA-WebFace*&31.539 &0.910 & 24.102  & 0.832 &0.049 &0.178 &0.064 & 0.295 &0.992 &0.937
 \\
			                        &VGGFace2 &31.322 &0.898 & 23.949    &0.831 &0.043  &0.162 &0.216 &0.648 &0.962 &0.841 \\
			
			\hline
	  \end{tabular}}
   
\label{table2}
   % \end{center}
\end{table*}

\begin{figure*}[t]
	\centering
	\includegraphics[width=160mm]{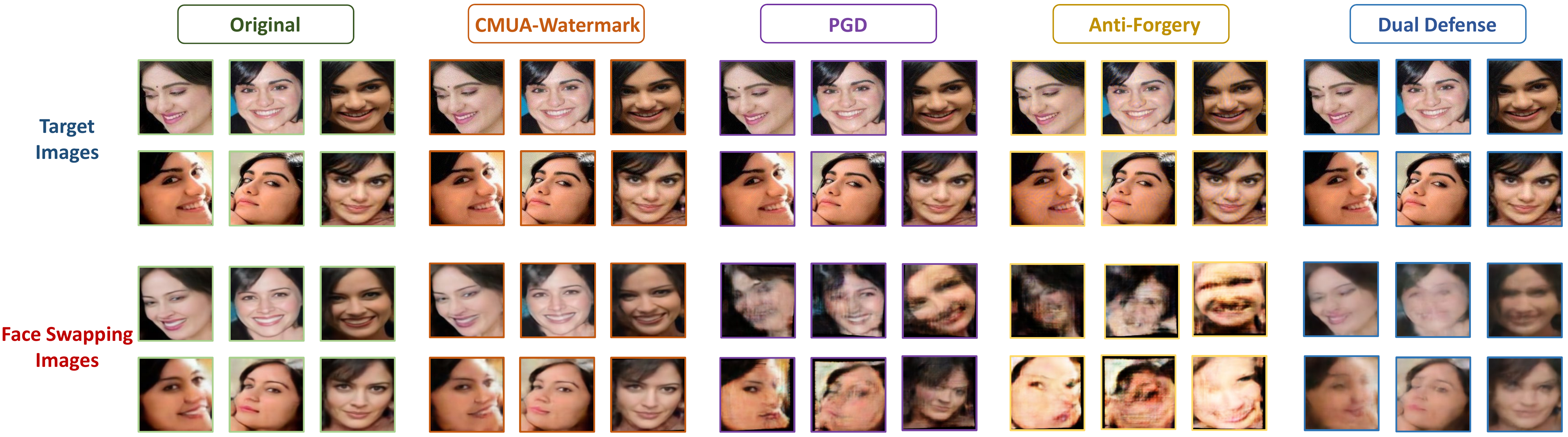}
         \caption{Visualization results of four methods on the VGGFace2 dataset. \emph{Target Images} in different methods indicate the generated adversarial examples (CMUA-Watermark, PGD and Anti-Forgery) or watermarked images (Dual Defense).}	
 \label{vm2}
\end{figure*} 

\begin{figure*}[t]
	\centering
	\includegraphics[width=160mm]{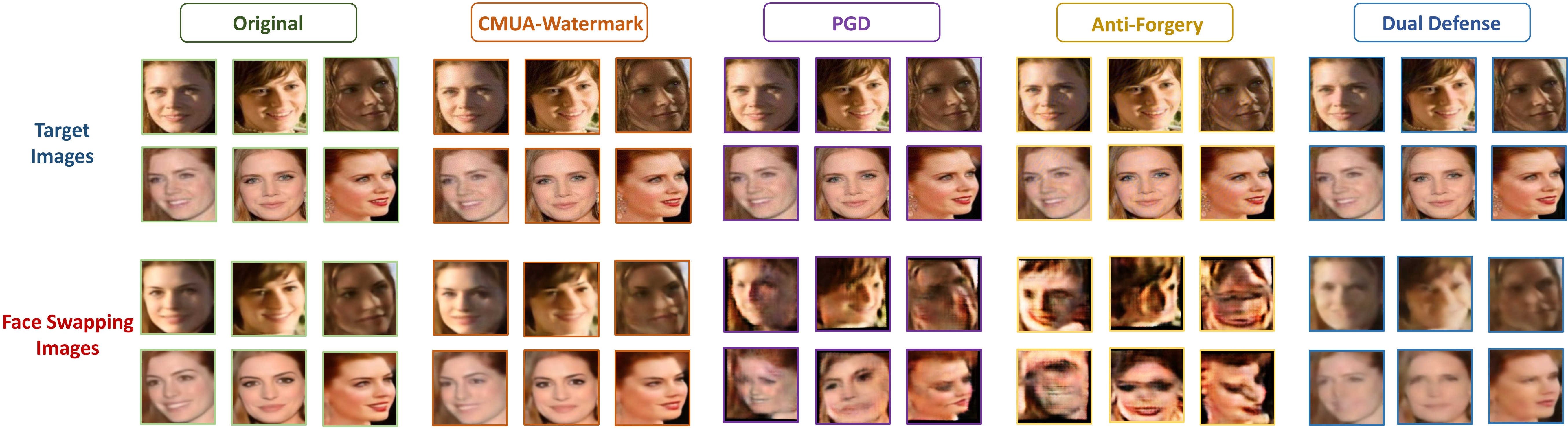}
         \caption{Visualization results of four methods on the Casia-WebFace dataset. \emph{Target Images} in different methods indicate the generated adversarial examples (CMUA-Watermark, PGD and Anti-Forgery) or watermarked images (Dual Defense).}	
 \label{casia}
\end{figure*} 

\begin{figure*}[t]
	\centering
	\subfloat[Gaussian Blur (kernel size=5)]{\includegraphics[width=55mm]{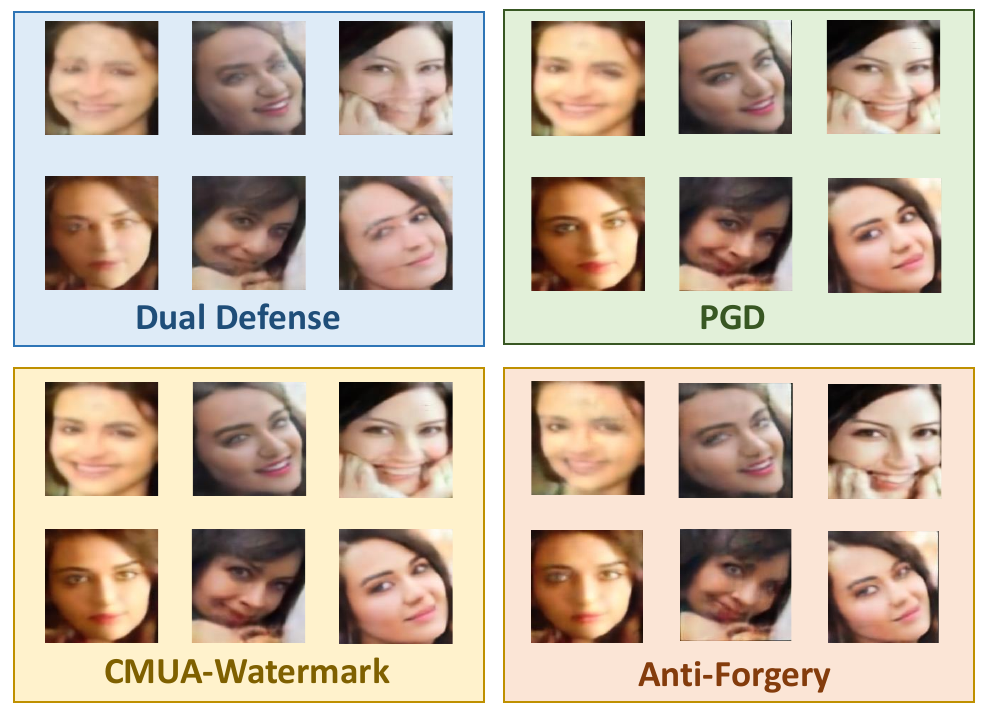}}\hspace{5mm}
        \subfloat[JPEG 50 (QF=50)]{\includegraphics[width=55mm]{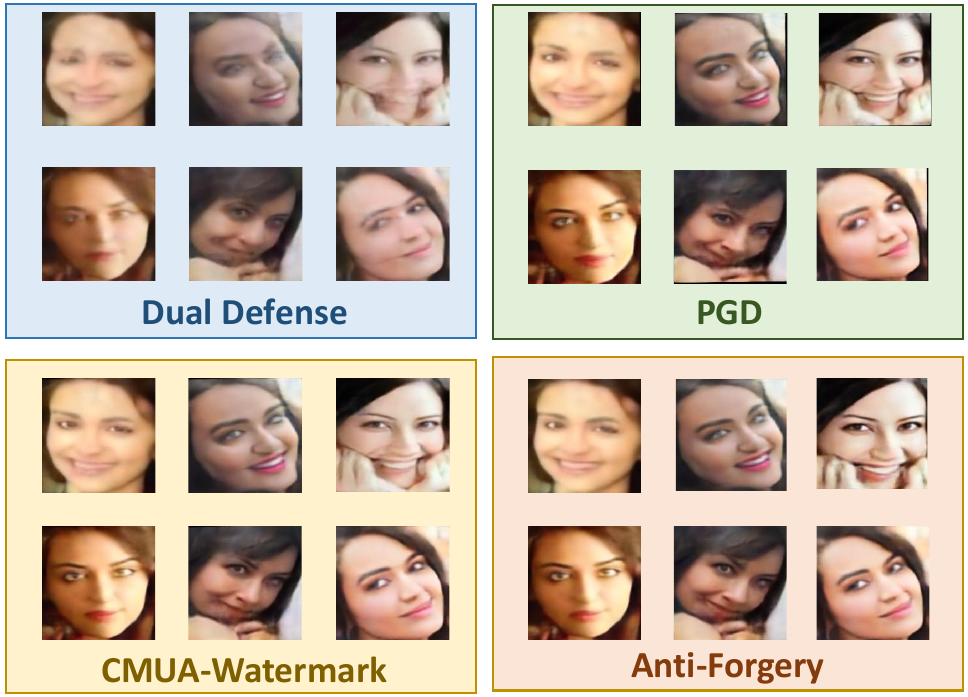}}\hspace{5mm}
        \subfloat[Resize (1/2)]{\includegraphics[width=55mm]{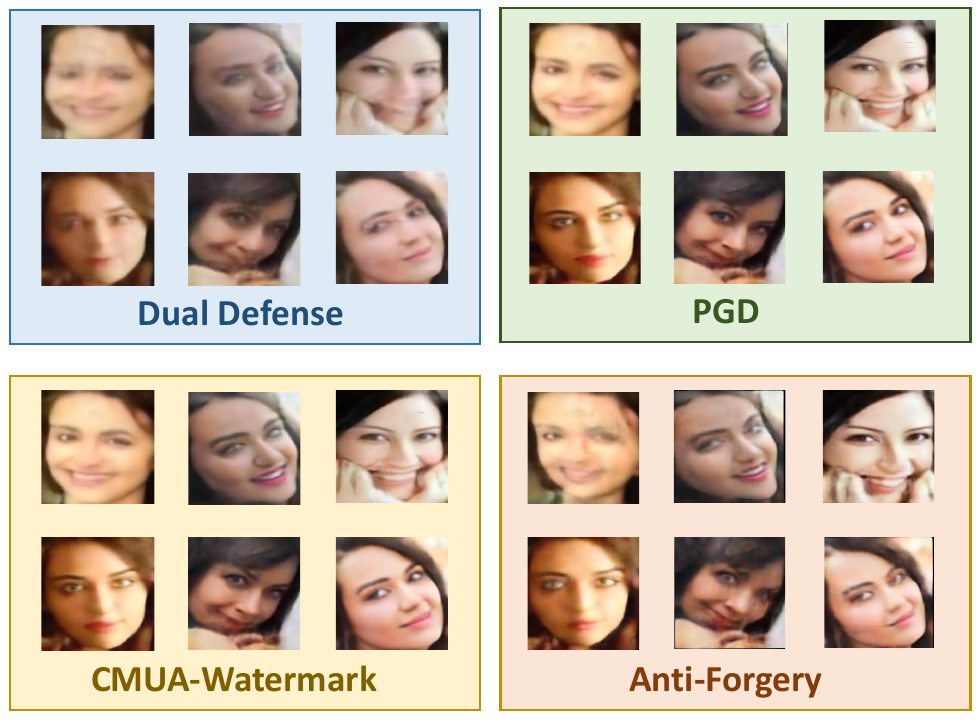}}\hspace{5mm}
         \caption{Visualization results of four methods under robust scenarios. }	
 \label{com3}
\end{figure*}

\begin{table}[t]
\caption{Results of Dual Defense in robust settings with various different image post-processing operations on VGGFace2 dataset.}
	% \renewcommand\arraystretch{1.2}  % 调整行高
	% \scriptsize  % 字号
 % \setlength\tabcolsep{2pt} 
\resizebox{\hsize}{!}{
	% %\resizebox{\linewidth}{1} % 这句总是报错
 % \scalebox{0.25} % dz换了一个表格缩放语句
   \begin{threeparttable}{
       \begin{tabular}{cccccccc}
     
       \toprule %添加表格头部粗线
       % \specialrule{0em}{3pt}{3pt}
		
            % \multirow{2}{*}{\textbf{Processing}}  &\multirow{2}{*}{\textbf{Parameters}}  & 
            % \\ \cline{3-5} \cline{6-8}
			Processing  &Parameters &$FN_{acc}$ $\downarrow$  & $SR_{mask}$  &$Acc_{org}$$\uparrow$ &$Acc_{adv}$$\uparrow$ \\
           \midrule

			%  \multicolumn{9}{c}{SST}\\
			
			\multirow{4}{*}{JPEG} &30 & 0.298  & 0.315& 0.927 &0.818\\
			&50 & 0.162 &0.408 & 0.952 &0.846 \\
			  &70& 0.091 &0.517 & 0.965 &0.857 \\
                &90& 0.022 &0.686 & 0.977 &0.882 \\
			\hline
                \multirow{4}{*}{\makecell{Gaussian \\ Noise}} &0.001 &0.082 &0.558 & 0.964 &0.876 \\
			&0.002 &0.058 &0.622 & 0.932 &0.850 \\
			  &0.003 & 0.027 &0.725 & 0.907 &0.830 \\
                &0.004 &0.007 &0.819 & 0.880 &0.806 \\
			\hline
              \multirow{4}{*}{Resize} &3/4 &0.018 &0.619 & 0.981 &0.902 \\
			&5/4 &0.004 &0.739 & 0.983 &0.920 \\
			  &3/2 & 0.067 &0.756 & 0.984 &0.920 \\
                &7/4 &0.045 &0.786 & 0.983 &0.920 \\
			\hline
                 \multirow{4}{*}{\makecell{Salt\&pepper\\ Noise}} &0.001 &0.025 &0.744 & 0.956 &0.864 \\
			&0.002 &0.001 &0.700 & 0.976 &0.893 \\
			  &0.003 & 0.011 & 0.678 & 0.992 &0.967 \\
                &0.004 &0.014 &0.692 & 0.961 &0.878 \\
		
   \bottomrule
    
	\end{tabular}}
  \end{threeparttable}}
 
	\label{ROBUST-v}

\end{table}

\begin{table}[t]
\caption{Results of Dual Defense in robust settings with various different image post-processing operations on CasiaWebFace dataset.}
	% \renewcommand\arraystretch{1.2}  % 调整行高
	% \scriptsize  % 字号
 % \setlength\tabcolsep{2pt} 
\resizebox{\hsize}{!}{
	% %\resizebox{\linewidth}{1} % 这句总是报错
 % \scalebox{0.25} % dz换了一个表格缩放语句
   \begin{threeparttable}{
       \begin{tabular}{cccccccc}
     
       \toprule %添加表格头部粗线
       % \specialrule{0em}{3pt}{3pt}
		
            % \multirow{2}{*}{\textbf{Processing}}  &\multirow{2}{*}{\textbf{Parameters}}  & 
            % \\ \cline{3-5} \cline{6-8}
			Processing  &Parameters &$FN_{acc}$ $\downarrow$ & $SR_{mask}$ &$Acc_{org}$$\uparrow$ &$Acc_{adv}$$\uparrow$ \\
           \midrule

			%  \multicolumn{9}{c}{SST}\\
			
	\multirow{4}{*}{JPEG} &30 & 0.280& 0.477 & 0.9849 &0.948  \\
			&50 & 0.132 &0.518 &0.993 &0.968\\
			  &70 & 0.062 &0.536 & 0.994 &0.974 \\
                &90 & 0.048 &0.590 & 0.997 &0.981\\
			\hline
                \multirow{4}{*}{\makecell{Gaussian \\ Noise}} &0.001& 0.002 &0.647 & 0.993 &0.963\\
			&0.002& 0.008 &0.659 & 0.993 &0.987\\
			  &0.003 & 0.010 &0.660 & 0.992 &0.979 \\
                &0.004 & 0.015 &0.676 & 0.989 &0.966 \\
			\hline
              \multirow{4}{*}{Resize} &3/4 & 0.161 &0.462& 0.996 &0.976\\
			&5/4 & 0.036 &0.523 & 0.998 &0.984\\
			  &3/2 & 0.062 &0.574 & 0.997 &0.983 \\
                &7/4 & 0.035 &0.528 & 0.999 &0.988 \\
			\hline
                 \multirow{4}{*}{\makecell{Salt\&pepper\\ Noise}} &0.001 & 0.008 &0.647 & 0.996 &0.980\\
			&0.002 &0.012 &0.659 & 0.995 &0.976\\
			  &0.003 &0.015  & 0.678 & 0.993 &0.969\\
                &0.004 &0.017 &0.672 & 0.993 &0.962\\
		
   \bottomrule
    
	\end{tabular}}
  \end{threeparttable}}
 
	\label{ROBUST-c}

\end{table}

\begin{table*}
 \caption{Comparison of Dual Defense and other active defense methods. The best results are emphasized in bold. \textbf{N/A} means that the method lacks this function and the data cannot be obtained.}
	% \renewcommand\arraystretch{2}
	% \newcommand{\tabincell}[2]
	% \footnotesize
	\centering
	
	\renewcommand\arraystretch{1.2}  % 调整行高
 \resizebox{\hsize}{!}{
    % \scalebox{0.5} % dz换了一个表格缩放语句

   % \begin{center}
       \begin{tabular}{cccccccccccccc}
			\hline
            %     &Method & Adversariality &Traceability\\
 		\multirow{2}{*}{\textbf{Dataset}} &\multirow{2}{*}{\textbf{Method}} & \multicolumn{3}{c}{\textbf{Original}}& \multicolumn{3}{c}{\textbf{JPEG ($QF=50$)}} & \multicolumn{3}{c}{\textbf{Gaussian Noise ($\sigma=0.005$)}}& \multicolumn{3}{c}{\textbf{Resize ($1/2$)}}\\
            \cmidrule(r){3-5}\cmidrule(r){6-8} \cmidrule(r){9-11} \cmidrule(r){12-14}
   
          & &$FN_{acc}$ $\downarrow$ &$Acc_{org}$$\uparrow$ &$Acc_{adv}$$\uparrow$ &$FN_{acc}$ $\downarrow$ &$Acc_{org}$$\uparrow$ &$Acc_{adv}$$\uparrow$&$FN_{acc}$ $\downarrow$ &$Acc_{org}$$\uparrow$ &$Acc_{adv}$$\uparrow$&$FN_{acc}$ $\downarrow$ &$Acc_{org}$$\uparrow$ &$Acc_{adv}$$\uparrow$\\
			\hline
			\multirow{4}{*}{\makecell{VGG-\\Face2}} &PGD~\cite{madry2017towards} &0.009  &N/A  &N/A    &0.396 &N/A &N/A  &0.024 &N/A &N/A &0.648 &N/A  &N/A \\
                                     & CMUA-Watermark~\cite{huang2022cmua} &0.188 &N/A &N/A     &0.612 &N/A &N/A   &0.036 &N/A &N/A   &0.668 &N/A  &N/A   \\
                                      & Anti-Forgery~\cite{wang2022anti} &\textbf{0.000} &N/A &N/A     &0.703 &N/A &N/A   &0.065 &N/A &N/A   &0.245 &N/A  &N/A   \\
			                    & FakeTagger~\cite{wang2021faketagger} &1.000 &0.9871 &0.8295     &0.964 &\textbf{0.956} &0.802 &0.748 &\textbf{0.968} &0.803 &0.696 &0.980  &0.810  \\
			&\textbf{Dual Defense (ours)} &0.003 &\textbf{0.996} &\textbf{0.923}     &\textbf{0.014} &0.962 &\textbf{0.873} &\textbf{0.004} &0.967 &\textbf{0.854} &\textbf{0.084}  &\textbf{0.987}  &\textbf{0.885}  \\
   \hline
   			\multirow{4}{*}{\makecell{CASIA-\\WebFace}} & PGD~\cite{madry2017towards} &0.075  &N/A  &N/A    &0.461 &N/A &N/A  &0.107 &N/A &N/A &0.584 &N/A  &N/A \\
               & CMUA-Watermark~\cite{huang2022cmua} &0.461 &N/A &N/A     &0.497 &N/A &N/A   &0.153 &N/A &N/A   &0.636 &N/A  &N/A   \\
               & Anti-Forgery~\cite{wang2022anti} &\textbf{0.003} &N/A &N/A     &0.830 &N/A &N/A   &0.182 &N/A &N/A   &0.555 &N/A  &N/A   \\
			& FakeTagger~\cite{wang2021faketagger} &0.784 &0.997 &0.9617     &0.874 &0.942 &0.7875 &0.408 &0.9576 &0.733 &0.815 &0.580  &0.571  \\
			&\textbf{Dual Defense (ours)} &0.009 &\textbf{0.998} &\textbf{0.986}     &\textbf{0.113} &\textbf{0.994} &\textbf{0.974} &\textbf{0.009} &\textbf{0.968} &\textbf{0.893} &\textbf{0.328}  &\textbf{0.997}  &\textbf{0.966}  \\
			\hline
	  \end{tabular}}
  
	\label{3}
   % \end{center}
\end{table*}

\subsection{Total Training Loss}
Dual Defense simultaneously trains the model end-to-end in the three aspects of watermark invisibility, watermarked image adversariality, and watermark traceability. The training process details of Dual Defense can be referred to as Algorithm~\ref{alg:TARA-Watermark}.The total optimization objective $\mathcal{L}_{total}$ is as follows:
\begin{eqnarray}\label{totalloss}
\mathcal{L}_{total}=\alpha\mathcal{L} _{img}+   \beta \mathcal{L} _{wm},
\end{eqnarray}
where $\alpha$ and $\beta$ are empirically set to 0.5, 2 respectively. We present the experiments and analysis used to determine the weights in the experiments section.

\section{Experiments}
\subsection{Experimental Settings}

\subsubsection{Datasets.} We use two large-scale face recognition datasets for model training and verification, VGGFace2~\cite{8373813} and CASIA-WebFace~\cite{YiLLL14a}. The image size is $160\times160$, the watermark message is 30 bits long, representing over 1 billion different AI fingerprints, which is sufficient for practical use. We train the FaceSwap and the corresponding Dual Defense watermarking model on pairs of person image sets. The training set, verification set, and test set of each character are divided according to the ratio of 0.6 : 0.2 : 0.2.

\subsubsection{Training Details.}Our Dual Defense is implemented by PyTorch and executed on NVIDIA RTX 3090. Due to computational resource constraints, all images are resized to $160\times 160.$ The entire training process spans 2500 epochs with a batch size of 16. We empirically adjust the Adam optimizer~\cite{kingma2014adam} with an initial learning rate of 0.00005 for stable training. Additionally, since watermarked images undergo watermark extraction through the FaceSwap model, which introduces high-intensity image distortion, there is no need to add an extra noise pool for robust training throughout the process. To ensure that the watermark decoder learns from a carrier that retains most of the image features, we set $deiter$ to 30. In other words, for the first 30 epochs, only the encoder is trained. Starting from the 31st epoch, the decoder is introduced for end-to-end training.

\subsubsection{Comparisons.} Since Dual Defense is the first active defense method that combines both adversariality and traceability, we compare it with other active defense methods based on adversarial examples, namely CMUA-Watermark~\cite{huang2022cmua} and Anti-Forgery~\cite{wang2022anti}, the classic adversarial attack method PGD~\cite{madry2017towards}, and the active defense method FakeTagger~\cite{wang2021faketagger} based on deep watermarking. We strictly adhere to the experimental setups outlined in the papers of the comparison methods and conduct white-box training on FaceSwap. We compared their adversariality and traceability separately. Notably, the aforementioned methods only possess either adversariality or traceability, while Dual Defense possesses both of these capabilities simultaneously.

\subsubsection{Metrics.} We use PSNR, SSIM~\cite{Wang_Bovik_Sheikh_Simoncelli_2004} to evaluate the quality of the watermarked image after encoding the watermark, use the accuracy rate $Acc_{org}$ and $Acc_{adv}$ to evaluate the watermark recovery accuracy of the watermarked image and the disrupted image respectively. $L_{1}$, LPIPS~\cite{zhang2018image}, PSNR, SSIM and $SR_{mask}$~\cite{huang2022cmua} are used to measure the quality of disrupted images to reflect the adversariality of methods, but they mainly measure the color, texture, and structural differences in the image, they are traditional metrics used for evaluating natural images. Dual Defense mainly aims to disrupt the replacement of target face identity information by FaceSwap, focusing on the protection of facial features. Therefore, we propose to use the face recognition model FaceNet~\cite{schroff2015facenet} to evaluate the adversariality. If FaceNet cannot recognize face swapping images as the source face identity, the disruption is considered successful. In order to exclude the effect of the FaceSwap on the FaceNet recognition accuracy, we propose the FaceNet recognition accuracy index:
% therefore, we also use the success rate metric $SR_{mask}$ proposed by CMUA to focus on the face area of the image to evaluate the adversarial performance of the method. 
\begin{eqnarray}\label{FN ACC}
FN_{acc}=\frac{FN_{acc}(X_{adv}^{W_{ID}}\rightarrow source)}{FN_{acc}(X_{s}^{(t)}\rightarrow source)},
\end{eqnarray}
where ${FN_{acc}(X_{adv}^{W_{ID}}\rightarrow source)}$ represents the recognition accuracy of FaceNet to recognize the disrupted image as the source face, ${FN_{acc}(X_{s}^{(t)}\rightarrow source)}$ represents the accuracy of FaceNet recognizing the original face swapping image as the source person. A smaller $FN_{acc}$ indicates that the face swapping images deviate more from the source face, indicating that Dual Defense has better adversariality.

\subsection{Results of Dual Defense in Original Settings}
In this subsection, we present the results of Dual Defense on the original settings without image post-processing. Table~\ref{table2} shows the results of Dual Defense on two datasets, covering white-box, cross-task, and cross-dataset scenarios. Due to the unsatisfactory performance of other comparative methods under white-box setting, we solely present the performance of our Dual Defense across the various settings mentioned above. In the white-box scenarios, Dual Defense consistently achieves the $FN_{acc}$ below 0.01 on both datasets, while the watermark recovery accuracy remains consistently above 0.9. Cross-task refers to testing Dual Defense on the FaceSwap model of another pair of unknown characters in the same dataset. Dual Defense maintains excellent performance in both cross-task and cross-dataset scenarios. The results validate the excellent cross-task universality and dataset generalization ability of Dual Defense, demonstrating that a single-trained Dual Defense is not limited to defending against face swapping task for a single identity but can be universally applied across different datasets and identities involved in face swapping tasks.
% \begin{figure}
% 	\centering
% 	\includegraphics[width=70mm]{./vis.pdf}
%          \caption{Visualization results of Dual Defense.}	
%  \label{quality}
% \end{figure} 
\subsection{Results of Dual Defense in Robust Settings}

In real channels and social networks, images often undergo various post-processing operations. Therefore, we evaluate the adversariality and traceability of Dual Defense against FaceSwap under four common image post-processing operations: JPEG compression, Gaussian noise, resizing and salt\&pepper noise. As shown in Table~\ref{ROBUST-v} and Table~\ref{ROBUST-c}, Dual Defense maintains excellent adversariality and traceability. Although high-intensity JPEG compression weakens the overall performance of Dual Defense, even at a JPEG compression factor of 30, the $FN_{acc}$ remains below 0.3, and the watermark recovery accuracy stays above 0.8. Considering all scenarios, when the watermarked image is subjected to various processing operations, the proposed Dual Defense consistently maintains excellent performance, thereby validating the feasibility of our method in practical scenarios.

\subsection{Comparison with Other Active Defense Methods}

We report the experimental results comparing Dual Defense to the other four methods in Table~\ref{3} and visualize their performance in original and robust scenarios in Fig.~\ref{vm2}, Fig.~\ref{casia} and Fig.~\ref{com3}. PGD, CMUA-Watermark, and Anti-Forgery are gradient-based adversarial attack methods without traceability, so traceability metrics are unavailable, denoted as $N/A$. We only compare their adversarial performance and contrast the traceability with FakeTagger. As shown in Table~\ref{3}, under the original setting, Anti-Forgery demonstrates the strongest adversariality due to its prolonged iterations, followed by Dual Defense. Image processing operations notably decrease the adversariality of the other three methods, while Dual Defense maintains exceptional adversariality, especially following JPEG compression, where its $FN_{acc}$ decreases by over 70$\%$ compared to Anti-Forgery. This observation implies that gradient attack-based adversarial example methods often lack robustness against FaceSwap, thereby limiting their practical applicability. Moreover, Dual Defense achieves consistently higher watermark recovery accuracy compared to FakeTagger while ensuring adversariality. Notably, on the CASIA dataset, when images are resized, Dual Defense's $Acc_{org}$ and $Acc_{adv}$ are approximately 40$\%$ higher than those of FakeTagger.

\subsection{Visual comparison and analysis}
In Fig.~\ref{vm2} and Fig.~\ref{casia}, we present the visual performance comparison between Dual Defense and the other three adversarial example methods on two datasets under the original setting. Notably, we exclude the visualization results of FakeTagger due to its lack of adversariality. As shown in Fig.~\ref{vm2} and Fig.~\ref{casia}, the adversarial examples generated by CMUA-Watermark, PGD, and Anti-Forgery exhibit clear and uniform noise patterns, especially Anti-Forgery, which contains noticeable noise. This is due to the extended iteration cycles for a single image, resulting in noticeable noise being introduced into the carrier, contributing to its high attack effectiveness. However, this characteristic is easily detectable and can be circumvented by malicious forgers. The watermarked images obtained by Dual Defense also show some quality degradation, but compared to other methods, they exhibit a more natural slight blurring rather than regular noise, which avoids raising suspicion from attackers. For disrupted images of CMUA-Watermark, minor changes in facial feature positions are observed while preserving the overall facial structure. The disrupted images produced by Anti-Forgery are the most severely damaged, but this is due to a significant reduction in the quality of its adversarial examples. In contrast, the watermarked images obtained by Dual Defense not only ensure the invisibility of the watermark but also completely blur the facial features, making them visually unrecognizable, thereby ensuring facial anonymity. 

Fig.~\ref{com3} displays the disrupted images obtained from the four methods under robust scenarios. Except for Dual Defense, the other three methods exhibit a noticeable decrease in adversariality after undergoing image processing operations, while Dual Defense still maintains excellent adversariality, blurs facial features, and conceals identity information. This indicates that Dual Defense possesses not only traceability robustness but also promising adversarial robustness.

\subsection{Comprehensive Defense Performance Evaluation}
In correspondence to the complex scenarios that facial images may encounter, as proposed in Section~\ref{PS}, we comprehensively assess the overall defense success rate of active defense methods from multiple perspectives. As shown in Table~\ref{acc}, adversarial example methods experience a significant reduction in adversarial performance after undergoing several common image post-processing steps, with the resulting perturbed images still displaying distinct source facial identity features, enabling correct recognition by facial recognition models. The introduced Dual Defense method offers traceability mechanisms in cases of weakened adversarial effectiveness, enhancing adversarial robustness while enabling image tracing. Therefore, in this section, we trace images recognized correctly by the FaceNet model in cases of adversarial attack failure to evaluate the comprehensive defense performance of Dual Defense. As shown in Table~\ref{acc}, across various complex scenarios, for images where attacks fail, Dual Defense consistently maintains a traceability accuracy of 0.9 or above. This demonstrates that in situations of poor adversarial effectiveness, Dual Defense can effectively provide auxiliary means to assist network administrators or the original target users in tracing the source.
\begin{table}
\caption{Comprehensive defense performance evaluation of Dual Defense in robust scenarios. The test images are all face-swapping images that can be successfully identified as the source face.}
	% \renewcommand\arraystretch{1.2}  % 调整行高
	% \scriptsize  % 字号
 % \setlength\tabcolsep{2pt} 
\resizebox{\hsize}{!}{
	% %\resizebox{\linewidth}{1} % 这句总是报错
 % \scalebox{0.25} % dz换了一个表格缩放语句
   \begin{threeparttable}{
       \begin{tabular}{cccccccc}
     
       \toprule %添加表格头部粗线
       % \specialrule{0em}{3pt}{3pt}
		
            \multirow{2}{*}{\textbf{Processing}}  &\multirow{2}{*}{\textbf{Parameters}}  & \multicolumn{2}{c}{\textbf{VGGFace2}} & \multicolumn{2}{c}{\textbf{CASIA-WebFace}}
            \\ \cline{3-4} \cline{5-6}
			&  &$Acc_{org}$$\uparrow$ &$Acc_{adv}$$\uparrow$ &$Acc_{org}$$\uparrow$ & $Acc_{adv}$$\uparrow$\\
           \midrule

			%  \multicolumn{9}{c}{SST}\\
			
			\multirow{4}{*}{JPEG} &30  & 0.945 &0.932 & 0.972 &0.975  \\
			&50 & 0.952 &0.913 & 0.993 &0.983\\
			  &70& 0.978 &0.919 & 0.994 &0.989 \\
                &90 & 0.982 &0.932 & 0.997 &0.991\\
			\hline
                \multirow{4}{*}{\makecell{Gaussian \\ Noise}} &0.001 & 0.972 &0.916 & 0.995 &0.976\\
			&0.002& 0.961 &0.925 & 0.993 &0.983\\
			  &0.003 & 0.957 &0.907& 0.992 &0.979 \\
                &0.004& 0.928 &0.914  & 0.995 &0.986\\
			\hline
              \multirow{4}{*}{Resize} &3/4 & 0.987 &0.932 & 0.998 &0.966\\
			&5/4 & 0.987 &0.946 & 0.998 &0.958\\
			  &3/2 & 0.988 &0.948 & 0.997 &0.973 \\
                &7/4 & 0.988 &0.946 & 0.999 &0.963\\
			\hline
                 \multirow{4}{*}{\makecell{Salt\&pepper\\ Noise}} &0.001 & 0.968 &0.914 & 0.997 &0.987\\
			&0.002 & 0.982 &0.935 & 0.998 &0.978\\
			  &0.003 & 0.992 &0.967 & 0.998 &0.976 \\
                &0.004 & 0.954 &0.918 & 0.997 &0.974\\
		
   \bottomrule
    
	\end{tabular}}
  \end{threeparttable}}
 
	\label{acc}

\end{table}

\subsection{Ablation Study}

In this section, we first  investigate the significance of the proposed original-domain facial feature attack method. Then, we discuss the effectiveness of the various contributions proposed. 

We introduce a watermark encoding strategy based on original-domain feature impersonation attack, using the watermarked image of the target face as the imitation target for targeted attacks. We analyze the drawbacks of two other attack methods and explain the rationale behind choosing the in-domain feature impersonation attack approach.

\begin{figure}
	\centering
	\subfloat[]{\includegraphics[width=41mm]{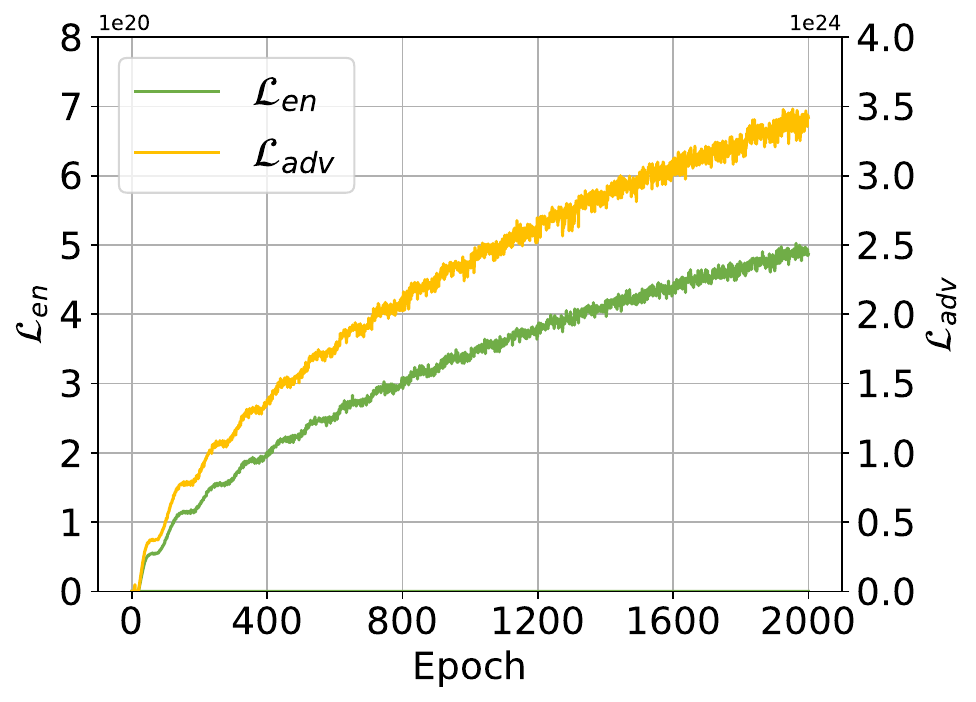}}
	\subfloat[]{\includegraphics[width=41mm]{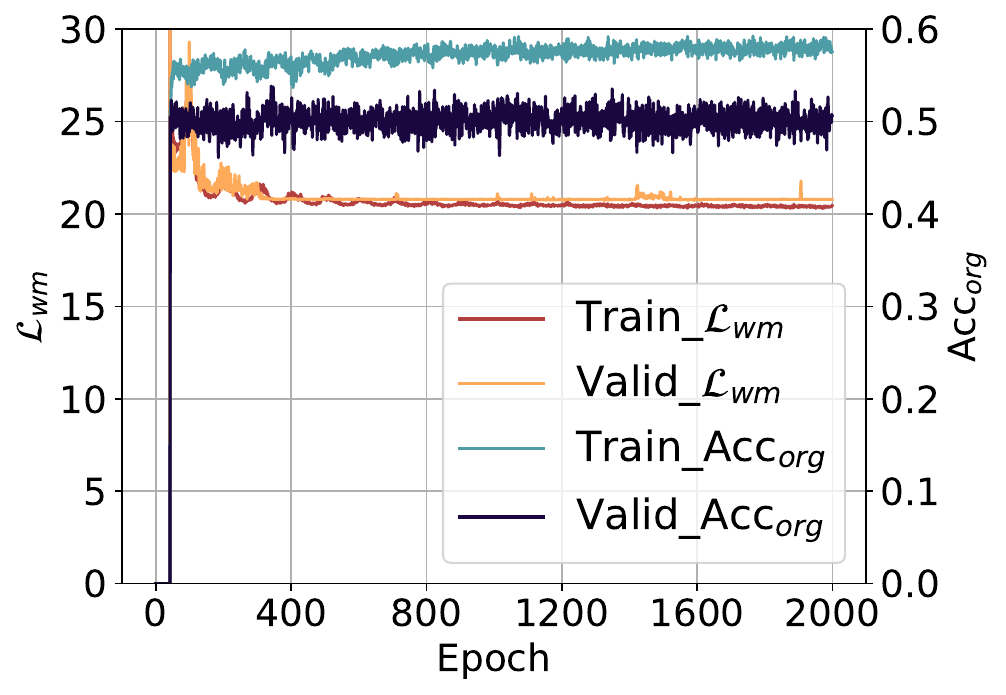}}
        \quad
	\subfloat[]{\includegraphics[width=41mm]{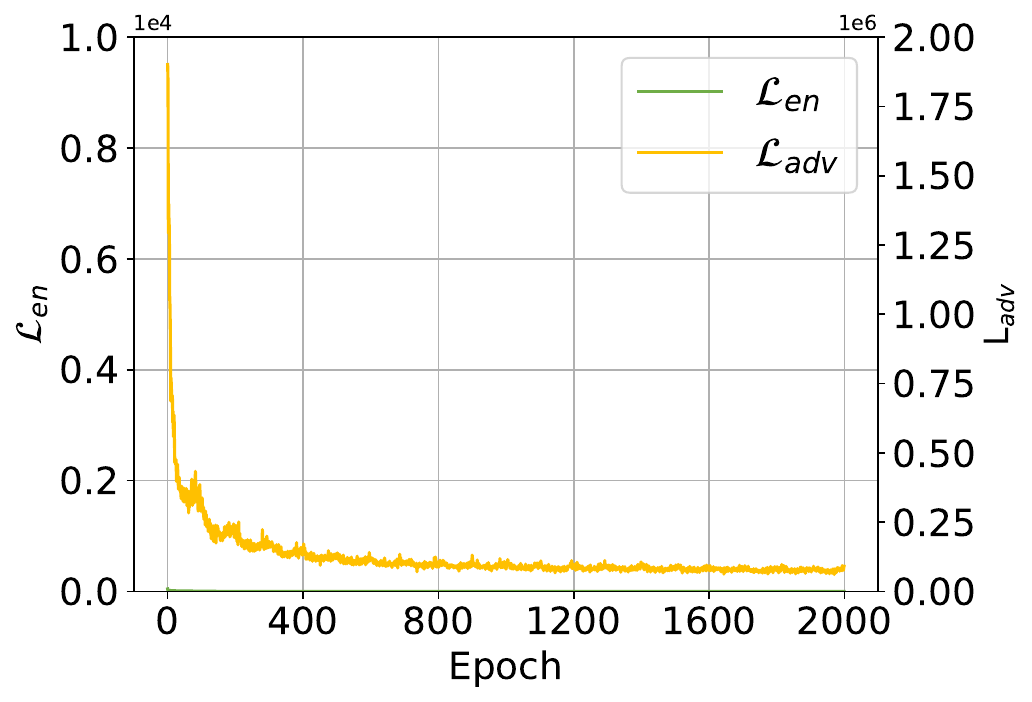}}
	\subfloat[]{\includegraphics[width=41mm]{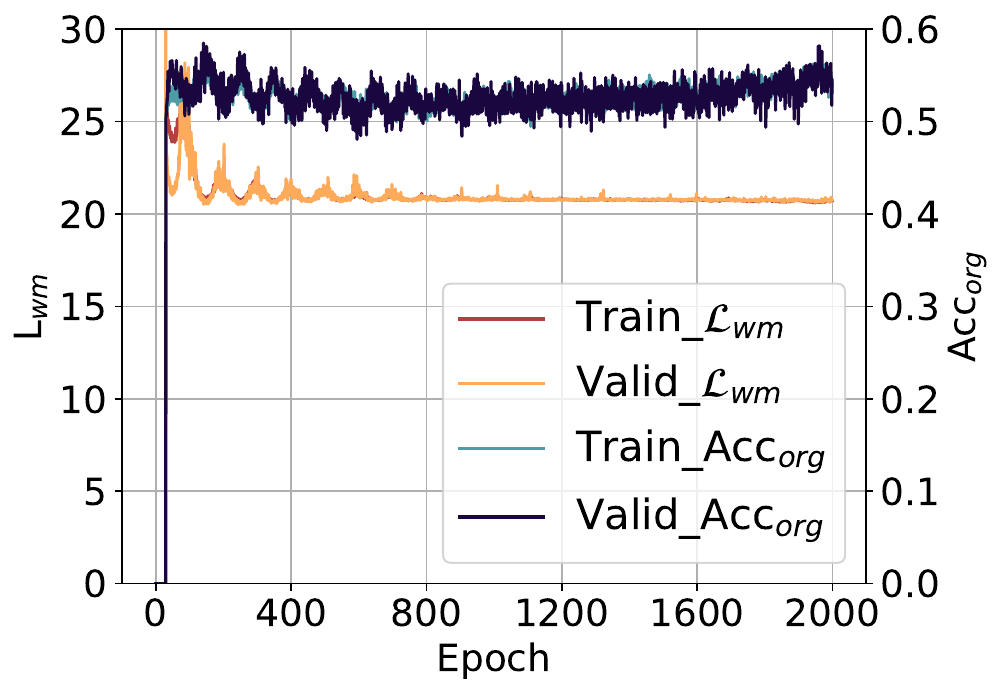}}
        \caption{Image loss during training and traceability loss, as well as watermark recovery accuracy during training and validation. (a-b) Untargeted attack. (c-d) Targeted attack using original carrier Images.}	
 \label{loss}
\end{figure}

\textbf{Untargeted attack} can be achieved by maximizing the distance between the faceswapped image of the watermarked image and the faceswapped image of the original carrier image (Eq.~\ref{un}). As depicted in Fig.~\ref{loss} (a-b), untargeted attack result in a loss of watermark image quality and an increase in adversarial loss. This is because the watermarked image obtained during the initial iterations loses almost all semantic and color features. Further disrupting with the faceswapped image prevents the optimization of watermarked image quality, leading to gradient explosion and the inability to extract the watermark message.

\begin{eqnarray}
\label{un}
\mathcal{L} =1-\frac{1}{N} \sum_{i=1}^{N} (FS(X_{t}^{W_{ID}},'{L_{s}}')-FS(X_{t},'{L_{s}}') )^{2} .
\end{eqnarray}

\textbf{Targeted attack based on the original carrier image} can be achieved by minimizing the reconstruction distance between the face swapped image of the watermark image and the reconstructed image of the original carrier (Eq.~\ref{oc}). As shown in Fig.~\ref{loss} (c-d), while the quality of the watermark image can continuously improve, there remains a significant difference in quality between the reconstructed watermark image and the original image reconstruction. This is still attributed to the quality degradation of the watermark image in the initial iterative steps, preventing progressive impersonation attacks and hindering watermark extraction.

\begin{eqnarray}
\label{oc}
\mathcal{L} =\frac{1}{N} \sum_{i=1}^{N} (FS(X_{t}^{W_{ID}},'{L_{s}}')-FS(X_{t},'{L_{t}}') )^{2} .
\end{eqnarray}

\begin{table}
	 \caption{Ablation study on adversarial optimization strategy. \emph{D} represents the discriminator, \emph{Adv} represents the original-domain facial feature attack method. The best results are emphasized in bold. }
	\renewcommand\arraystretch{1.2}  % 调整行高
	% \scriptsize  % 字号
 % \setlength\tabcolsep{2pt} 
\resizebox{\hsize}{!}{
	% %\resizebox{\linewidth}{1} % 这句总是报错
 % \scalebox{0.25} % dz换了一个表格缩放语句
 
       \begin{tabular}{cccccccccc}
       % \specialrule{0em}{3pt}{3pt}
			\hline
           \multirow{2}{*}{\textbf{Dataset}}& \multirow{2}{*}{\textbf{Method}} & \multicolumn{2}{c}{\textbf{Invisibility}} & \multicolumn{2}{c}{\textbf{Adversariality}}& \multicolumn{2}{c}{\textbf{Traceability}}
            \\ \cline{3-4} \cline{5-6}\cline{7-8}
			& &PSNR $\uparrow$ &SSIM$\uparrow$ & $FN_{acc}$ $\downarrow$& $SR_{mask}$ $\uparrow$ &$Acc_{org}$$\uparrow$ & $Acc_{adv}$$\uparrow$\\

			\hline
			%  \multicolumn{9}{c}{SST}\\
			
			\multirow{4}{*}{\makecell{VGG-\\Face2}} &w/o SIC& 27.981 & 0.855 &\textbf{0.000} & 0.742 & 0.839 &0.802  \\
			 &w/o SENet& 31.361 &0.914 &0.012 & 0.714 & 0.978 &0.912\\
			 &w/o SIC$+$SENet & 27.412 & 0.809 &0.196 & 0.523 & 0.838 &0.812 \\
                &w/o Adv& \textbf{42.112} & \textbf{0.996} &0.990 & 0.132 & 0.989 &\textbf{0.988}\\
                &Dual Defense& 31.782 & 0.917 &0.003 &\textbf{0.772} & \textbf{0.996} &0.923\\
			\hline
             \multirow{4}{*}{\makecell{CASIA-\\WebFace}} &w/o SIC & 30.213 & 0.859 &\textbf{0.006} & 0.457 & 0.978 &0.968  \\
			 &w/o SENet& 31.824 & 0.936 &0.0136 & 0.537 & 0.985 &0.976\\
			 &w/o SIC$+$SENet & 27.983 & 0.789 &0.363 & 0.359 & 0.962 &0.911 \\
                &w/o Adv& \textbf{41.036} & \textbf{0.995} &0.997 & 0.218 & 0.993 &0.986\\
                &Dual Defense&31.830 & 0.925 &0.009 & \textbf{0.683} & \textbf{0.998} &\textbf{0.986}\\
			\hline
		
	\end{tabular}}

	\label{Ablation}

\end{table}

\begin{table}
	% \renewcommand\arraystretch{2}
	% \newcommand{\tabincell}[2]
	% \footnotesize
  \caption{Quantitative results of Dual Defense under different watermark lengths. The training and test dataset is CASIA-WebFace. The best results are emphasized in bold.}
	\centering

 %        \setlength\tabcolsep{5pt} 
	% \renewcommand\arraystretch{1.2}  % 调整行高
    % \scalebox{0.88} % dz换了一个表格缩放语句
\resizebox{\hsize}{!}{
   % \begin{center}
       \begin{tabular}{cccccccc}
			\hline
            %     &Method & Adversariality &Traceability\\
 		\multirow{2}{*}{\textbf{Length}} & \multicolumn{2}{c}{\textbf{Invisibility}}& \multicolumn{2}{c}{\textbf{Adversariality}}  & \multicolumn{2}{c}{\textbf{Traceability}}\\
            \cmidrule(r){2-3}\cmidrule(r){4-5} \cmidrule(r){6-7} 
   
           &PSNR$\uparrow$ & SSIM$\uparrow$ & $FN_{acc}$ $\downarrow$& $SR_{mask}$ $\uparrow$&$Acc_{org}$$\uparrow$ &$Acc_{adv}$$\uparrow$\\
           \Xcline{1-1}{0.4pt}
			\hline
   
			 15 bits &31.217 &0.898  &0.011 &\textbf{0.671} &\textbf{0.999} &0.974 \\
                                        30 bits &\textbf{31.830} &\textbf{0.925} &\textbf{0.009}&0.683 &0.998 &\textbf{0.986} \\
			                         45 bits &30.736 &0.898 &0.019 &0.386 &0.998 &\textbf{0.985}\\
			\hline
	  \end{tabular}}

\label{bits}
   % \end{center}
\end{table}
\label{weightsapp}

\begin{table*}
	% \renewcommand\arraystretch{2}
	% \newcommand{\tabincell}[2]
	% \footnotesize
  \caption{Quantitative results of Dual Defense under different weights. The results of $\alpha \colon \beta$ were obtained with $\lambda _{en} \colon \lambda _{G} \colon \lambda _{adv}$ set to $0.8\colon0.1\colon0.1$, and the results of $\lambda _{en} \colon \lambda _{G} \colon \lambda _{adv}$ were obtained with $\alpha \colon \beta$ set to $0.5\colon2$. The training and test dataset is CASIA-WebFace. The best results are emphasized in bold.}
	\centering
   
 %        \setlength\tabcolsep{5pt} 
	% \renewcommand\arraystretch{1.2}  % 调整行高
    % \scalebox{0.88} % dz换了一个表格缩放语句
\resizebox{\hsize}{!}{
   % \begin{center}
       \begin{tabular}{ccccccccccccc}
			\hline
            %     &Method & Adversariality &Traceability\\
 		\multirow{2}{*}{\textbf{Weight}}&\multirow{2}{*}{\textbf{Ratio}} & \multicolumn{2}{c}{\textbf{Invisibility}}& \multicolumn{6}{c}{\textbf{Adversariality}}  & \multicolumn{2}{c}{\textbf{Traceability}}\\
            \cmidrule(r){3-4}\cmidrule(r){5-10} \cmidrule(r){11-12} 
   
           & &PSNR$\uparrow$ & SSIM$\uparrow$  &PSNR$\downarrow$ & SSIM$\downarrow$ & $L_{1}$$\uparrow$ & LPIPS$\uparrow$ & $FN_{acc}$ $\downarrow$& $SR_{mask}$ $\uparrow$&$Acc_{org}$$\uparrow$ &$Acc_{adv}$$\uparrow$\\
           \Xcline{1-1}{0.4pt}
			\hline
   
			\multirow{3}{*}{\makecell{$\alpha$$\colon$$\beta$\\ (0.8$\colon$0.1$\colon$0.1)}} &0.5$\colon$1 &32.404 &0.911  &23.341  &0.819 &0.055  &0.212 &0.011 &0.411 &0.993 &0.974 \\
                                        &0.5$\colon$2 &31.830 &0.925 & \textbf{22.363}    &\textbf{0.764} &\textbf{0.062} & \textbf{0.273} &0.009&\textbf{0.683} &\textbf{0.998} &0.986 \\
			                         &0.5$\colon$3 &31.925 &0.908 &23.193     &0.813  &0.055 &0.219 &0.012 &0.492 &0.998 &\textbf{0.992}\\
		
			\hline
   			\multirow{6}{*}{\makecell{$\lambda _{en}$$\colon$$\lambda _{G}$$\colon$$\lambda _{adv}$ \\ (0.5$\colon$2)}} &0.9$\colon$0.1$\colon$0.1 &32.295  &0.903  & 23.304  &0.819 &0.054  &0.216 &0.010 &0.470   &0.996 &0.985 \\
                                       &0.7$\colon$0.1$\colon$0.1 &31.354 &0.891 & 22.782  & 0.806 &0.058 &0.232 &\textbf{0.002} & 0.455 &0.998 &0.985
 \\

			                        &0.8$\colon$0.01$\colon$0.1 &31.841 &0.895 & 22.972    &0.811 &0.057  &0.222 &0.006 &0.463 &0.997 &0.986 \\
                           &0.8$\colon$0.2$\colon$0.1 &31.714 &0.894 & 22.749    &0.803 &0.058  &0.230 &0.004 &0.494 &0.998 &0.987\\
			
                    &0.8$\colon$0.1$\colon$0.01 &\textbf{36.19}5 &\textbf{0.956} & 24.585    &0.844 &0.037  &0.099 &0.755 &0.424 &0.997 &0.952 \\
                           &0.8$\colon$0.1$\colon$0.2 &29.160 &0.849 & 23.378    &0.818 &0.050  &0.187 &0.175 &0.408 &0.995 &0.922 \\
			\hline
	  \end{tabular}}

\label{cross}
   % \end{center}
\end{table*}
The original-domain facial feature attack of the watermarked image ensures a consistent gradient direction during the multi-task learning process. It reduces the quality difference between the watermarked image and the image to be imitated from the early iterations of the attack. This guarantees the watermarked image can effectively disrupt FaceSwap while optimizing the watermark decoder efficiently.

In addition, the watermark encoding network of Dual Defense incorporates SENet to guide watermark embedding and utilizes structural information compensation to enhance the quality of watermarked image reconstruction. We conducted ablation experiments to observe their impact on image invisibility, adversariality, and traceability. As shown in Table~\ref{Ablation}, SENet enhances the adversariality and traceability of Dual Defense but reduces the watermark invisibility. Image structure information compensation effectively enhances image quality but does not contribute to other aspects of performance. Dual Defense combines these two mechanisms to achieve effective trade-off in watermark performance. Additionally, Table~\ref{Ablation} indicates that the adversarial optimization method provides fundamental adversariality for Dual Defense.

We investigate the impact of watermark length on the performance of Dual Defense. As shown in Table~\ref{bits}, longer watermark length may cause a decrease in $SR_{mask}$, $FN_{acc}$ remains below 0.02, indicating the effectiveness of our method in identity protection. Under the three watermark lengths, the watermark recovery accuracy remains above 0.97. This experiment confirms that Dual Defense maintains outstanding performance across various watermark requirements. Since the watermark length of 30 bits is sufficient to meet the needs of real social networks, and has better adversariality, we choose 30 bits watermark for model training in our scheme.

 % $SR_{mask}$ is computed using the mask to locate facial regions for $L_{2}$ calculation, wherein facial feature deviation has a greater influence on its value compared to facial feature distortion, which does not fully explain the ability of the active defense method to protect identity information. 

We investigate the impact of different weight configurations in the loss function on the performance of our method. We utilize repeated cross-tests to identify the optimal weight configuration that achieves a balanced trade-off between various performance aspects. The experimental results are presented in Table~\ref{cross}. When setting $\beta$ to 1, we observe that the invisibility of the watermark reached its highest level. However, this came at the cost of compromised adversariality and traceability. On the other hand, as we increased the value of $\beta$, the accuracy of watermark recovery improved, but the adversariality also decreased. To strike a balance between adversariality and traceability, we selected a ratio of 0.5 : 2 for $\alpha$ to $\beta$. Subsequently, we determined the optimal internal image loss weights individually. Notably, when $\lambda _{en}\colon\lambda _{G}\colon\lambda _{adv}$  was set to 0.8 : 0.1 : 0.1, the overall performance of watermarking reached its optimum. 

\section{Conclusion}

% In this paper, we propose the first dual-effect active defense method that is both adversarial and traceable based on invisible deep watermarking. The proposed method can effectively prevent malicious face swapping in various social network scenarios, while enabling watermark extraction from disrupted images for identity traceability. or provide an auxiliary defense method for traceability in complex environments that are unfavorable to attacks. We introduce a watermark embedding method based on original domain feature impersonation attack, learn the robust adversarial features of the carrier, embed the watermark into the deep feature map, and effectively balance the multiple properties of the watermark through a GAN-based perceptual coding strategy. Experimental results demonstrate the dual-effect active defense capabilities of the method incorporating adversariality and traceability, along with its outstanding task and dataset generalization ability, and robustness to image post-processing operations, making it well-suited for practical applications.

In this paper, we propose the first dual-effect active defense method that is both adversarial and traceable based on invisible deep watermarking. The proposed method embeds a single invisible adversarial watermark for copyright tracking in social network face images. It can handle sudden malicious face-swaps during image propagation, disrupting the output of the face swapping model while ensuring the integrity of the imperceptible watermark. Authorized users can extract the watermark for traceability at any stage of image propagation. We introduce an adversarial watermark network based on the original-domain feature impersonation attack, which encodes robust adversarial features of the carrier image by learning the reconstruction features of the target face watermarked image. The watermark is optimally embedded into the deep feature maps of the carrier using a GAN-based perceptual encoding strategy, effectively balancing multiple watermark performance metrics during end-to-end optimization, resolving multi-objective optimization conflicts. Experimental results demonstrate the dual-effect active defense capabilities of the method incorporating adversariality and traceability, along with its outstanding task and dataset generalization ability, and robustness to image post-processing operations, making it well-suited for practical applications.

% In this paper, we propose the first dual-effect active defense method that is both adversarial and traceable based on invisible deep watermarking. The proposed method effectively thwarts malicious face swapping in various social network scenarios, while enabling watermark extraction from disrupted images for identity traceability. We first introduce an original-domain facial feature attack method for watermarked images, rendering the invisible robust watermark adversarial against FaceSwap. Then we employ a GAN-based perceptual adversarial encoding strategy to effectively balance the multiple performance of the watermark. Experimental results demonstrate the dual-effect active defense capabilities of the method incorporating adversariality and traceability, along with its outstanding task and dataset generalization ability, and robustness to image post-processing operations, making it well-suited for practical applications.

% \section*{Acknowledgments}
% This should be a simple paragraph before the References to thank those individuals and institutions who have supported your work on this article.

\bibliographystyle{IEEEtran}

\bibliography{tifs}

\vfill

\end{document}